\documentclass{article} % For LaTeX2e
\usepackage{iclr2027_conference,times}

\usepackage{amsmath,amsfonts,bm}

\def\eqref#1{equation~\ref{#1}}
\def\1{\bm{1}}

\DeclareMathAlphabet{\mathsfit}{\encodingdefault}{\sfdefault}{m}{sl}
\SetMathAlphabet{\mathsfit}{bold}{\encodingdefault}{\sfdefault}{bx}{n}

\usepackage{hyperref}
\usepackage{url}
\usepackage{algorithm}
\usepackage{algorithmic}
\usepackage{newfloat}
\usepackage{epsfig}
\usepackage{listings}
\usepackage{amsmath}      % 支持公式环境
\usepackage{amssymb}      % 支持更多数学符号，如 \mathbb
\usepackage{graphicx}     % 如果要插图
\usepackage{bm}           % 支持粗体数学符号
\usepackage{mathtools}    % math 优化和扩展（可选）
\usepackage{fontawesome}
\usepackage{booktabs}
\usepackage{multirow} 
\usepackage{xcolor}
\usepackage{placeins} 
\usepackage{colortbl} 
\usepackage{arydshln}
\usepackage{tabularx}
\usepackage{pifont}
\usepackage{makecell}
\usepackage{array}
\usepackage{tcolorbox}
\usepackage[table]{xcolor}
\usepackage{enumitem}     % 用于定制列表样式
\usepackage{colortbl}  % 用于表格颜色
\usepackage{caption}
\usepackage{graphicx}
\usepackage{adjustbox}
\usepackage{subcaption}
\usepackage{capt-of}

\title{ArmorOCR: Grounded Adversarial Visual Perception via Observation-Transferred Self-Distillation}

\author{
Linhan Cao$^{1,2}$\thanks{Equal contribution.
\quad $\dagger$Corresponding authors.},
Siyuan Li$^{1}$\footnotemark[1],
Jun Lan$^{1,\dagger}$,
Liangbo He$^{1}$,
Guannan Li$^{1}$,
Xiaolei Huang$^{1}$ \\
\textbf{
Jun Jia$^{2}$,
Shuheng Zhou$^{1}$,
Huijia Zhu$^{1}$,
Weiqiang Wang$^{1}$,
Wei Sun$^{3,\dagger}$
} \\
$^{1}$Ant Group,
$^{2}$Shanghai Jiao Tong University,
$^{3}$East China Normal University
}
\iclrfinalcopy % Uncomment for camera-ready version, but NOT for submission.
\begin{document}

\maketitle

\begin{abstract}
Large multimodal models (LMMs) have demonstrated strong OCR recognition capabilities, yet remain vulnerable to adversarial visual text that is readable to humans but challenging for models to localize and recognize. Existing OCR benchmarks mainly focus on natural or document-style text, while adversarial OCR evaluations remain limited in scale, task coverage, or region-aware evaluation. In this paper, we formulate adversarial OCR as a \textbf{grounded OCR perception} task and introduce \textbf{AdvSpot}, the first benchmark for grounded adversarial OCR evaluation.
AdvSpot comprises 383 images with region-level annotations, spanning 5 primary categories and 13 fine-grained adversarial OCR types. To address this challenge, we propose \textbf{ArmorOCR}, a two-stage training framework for robust adversarial OCR perception. ArmorOCR first acquires missing adversarial OCR perception from privileged transformed observations through On-Policy Self-Distillation (OPSD), and then refines grounded OCR perception through Group Relative Policy Optimization (GRPO) with task-conditioned rewards for localization, recognition, full spotting, and visual question answering (VQA). Experiments on our AdvSpot, other adversarial OCR benchmarks, and general OCR benchmarks demonstrate that ArmorOCR consistently improves adversarial OCR perception while preserving competitive general OCR capability. The benchmark and code are available at \url{https://github.com/ant-research/ArmorOCR}.
\end{abstract}

\vspace{-10pt}
\begin{figure}[h]
\centering
\centerline{\epsfig{figure=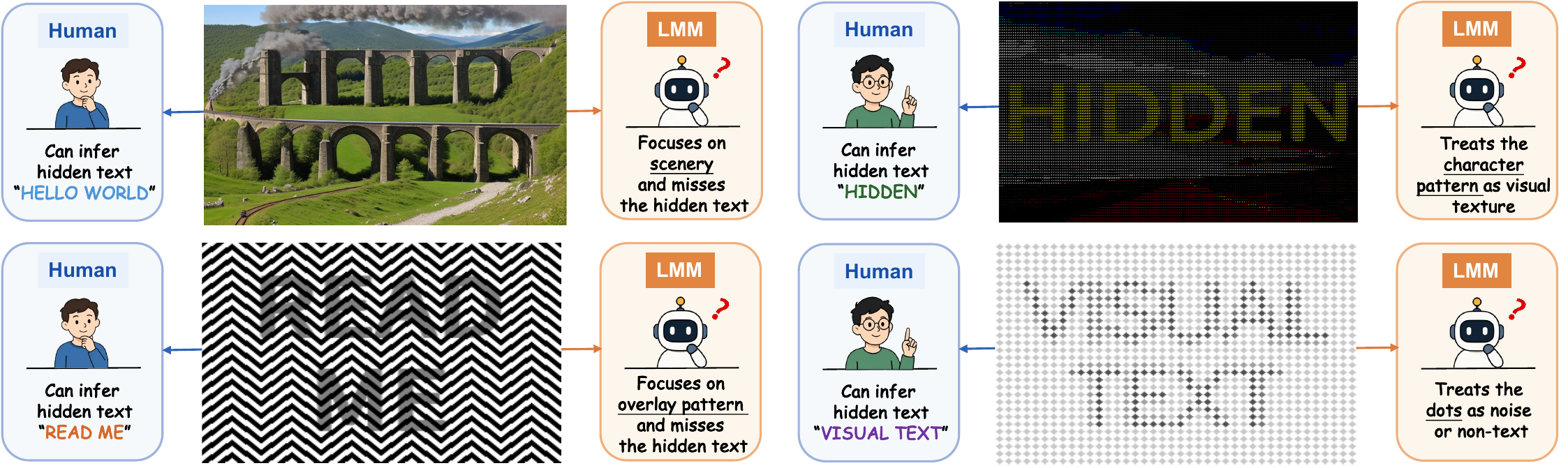,width=14cm}}
\caption{Examples of adversarial OCR patterns that reveal a human--AI perception gap: humans can easily recover the intended text from visual context, while current LMMs may mislocalize, overlook, or misread it.}
\label{fig:motivation}
\end{figure}
\vspace{-10pt}

\section{Introduction}

% 提出对抗ocr概念+研究意义
OCR is a fundamental capability of large multimodal models (LMMs)~\citep{zhang2023internlm,li2024llava,zhu2025internvl3}, yet its reliability under challenging visual conditions remains limited~\citep{li2025semvink,gao2025pixels}. As illustrated in Figure~\ref{fig:motivation}, certain visual text patterns expose a clear human--AI perception gap: \textbf{humans can easily recover the intended text from visual context, whereas current LMMs may mislocalize, overlook, or misread it.} We refer to such patterns as \emph{adversarial OCR patterns}. These patterns expose systematic weaknesses in LMMs' visual text perception, underscoring the need for systematic evaluation and targeted model improvement to bridge this gap and develop more 
robust OCR capabilities.

% 现在bench的不足
Existing OCR benchmarks~\citep{liu2024ocrbench,yang2025cc,fu2026ocrbench} have advanced general OCR evaluation across diverse natural visual content, but provide limited coverage of adversarial OCR patterns. Recent adversarial OCR benchmarks~\citep{xu2026vacot,li2026making} begin to address this setting, but remain limited in scale, task diversity, or region-level annotations. Moreover, recognition-only evaluation provides limited insight into whether a model has identified the relevant text region or merely inferred the answer from global context or linguistic priors.

% 我们的定义
% Motivated by this limitation, we formulate adversarial OCR perception as a \textbf{locate--read--understand} task, requiring models to localize adversarial text regions, recognize their content, and understand how adversarial OCR patterns affect text perception within their visual context. This region-grounded formulation makes model predictions verifiable and enables fine-grained diagnosis of adversarial OCR failures.

% Motivated by this limitation, we formulate adversarial OCR perception as a \textbf{grounded OCR perception} task, requiring models to localize adversarial text regions and robustly perceive their content under challenging visual conditions. This region-grounded formulation makes model predictions verifiable and enables fine-grained diagnosis of adversarial OCR failures.

Motivated by this limitation, we formulate adversarial OCR as a \textbf{grounded OCR perception} task, where models are expected to identify relevant text regions, recognize adversarial text content, and answer region-specific questions under challenging visual conditions. 
% This region-grounded formulation makes model predictions verifiable and enables fine-grained diagnosis of adversarial OCR failures.
This formulation supports region-level verification and more fine-grained analysis of adversarial OCR failures.

% 介绍我们的bench
% To instantiate this formulation, we introduce \textbf{AdvSpot}, the first benchmark for grounded adversarial OCR perception. AdvSpot provides the most comprehensive perception taxonomy of adversarial OCR patterns to date, covering 5 primary categories and 13 fine-grained types organized according to their underlying OCR failure mechanisms. It contains 379 images with region-level annotations, including bounding boxes, text transcriptions, category labels, and grounded question--answer pairs. Together, these annotations enable systematic evaluation of the locate--read--understand capability.

To instantiate this formulation, we introduce \textbf{AdvSpot}, the first benchmark for grounded adversarial OCR perception. AdvSpot provides the most comprehensive perception taxonomy of adversarial OCR patterns to date, covering 5 primary categories and 13 fine-grained types organized according to their underlying OCR failure mechanisms. It contains 383 images with region-level annotations, including bounding boxes, text transcriptions, category labels, and grounded visual question answering (VQA) pairs. Together, these annotations enable systematic evaluation of grounded adversarial OCR perception with respect to localization, recognition, and region-specific question answering.

Recent approaches~\citep{xu2026vacot,li2025semvink} to adversarial OCR perception rely on ``thinking-with-images'' strategies~\citep{hong2025deepeyesv2,song2025codedance}, applying transformations such as cropping, resizing, or flipping at inference time to reveal adversarial text that is difficult to perceive from the original view. Although effective, these approaches introduce additional inference latency and deployment complexity, motivating us to ask: \emph{Can the perceptual information revealed by transformed views be internalized into model parameters during training?}

To answer this question, we propose \textbf{ArmorOCR}, a two-stage training framework that internalizes transformation-revealed visual evidence for single-pass grounded adversarial OCR perception, without requiring additional visual transformations or tools at inference time. In Stage~1, we perform On-Policy Self-Distillation (OPSD) with response-region-aware token weighting: the student observes only the original image, while a teacher with the same backbone is conditioned on privileged transformed views. Token-level distribution guidance along the student's on-policy trajectories transfers the perception revealed by privileged observations to the student, establishing a robust perceptual foundation. However, self-distillation remains constrained by the teacher's performance ceiling and lacks explicit optimization of diverse grounded OCR objectives. Stage~2 therefore employs Group Relative Policy Optimization (GRPO) with task-conditioned rewards for localization, recognition, full spotting, and grounded VQA to jointly optimize these complementary capabilities.

Our contributions are summarized as follows:
\begin{itemize}

    \item We introduce \textbf{AdvSpot}, the first benchmark for grounded adversarial OCR perception, with the most comprehensive taxonomy of adversarial OCR patterns to date and region-level annotations enabling fine-grained evaluation of adversarial OCR perception.

    % \vspace{-5pt}

    \item We propose \textbf{ArmorOCR}, a two-stage framework that internalizes transformation-revealed perception through privileged observation transfer and refines grounded OCR perception with task-conditioned GRPO.

    % \vspace{-5pt}

    \item Extensive experiments on AdvSpot, existing adversarial OCR benchmarks, and general OCR benchmarks demonstrate that ArmorOCR consistently improves adversarial OCR perception while preserving competitive general OCR capability.

\end{itemize}

\vspace{-10pt}

\section{Related Work}

\subsection{OCR Perception Benchmarks}

% Existing OCR benchmarks~\citep{singh2019towards,mathew2021docvqa,masry2022chartqa,wang2024charxiv}, such as OCRBench~\citep{liu2024ocrbench}, CCOCR~\citep{yang2025cc}, and OCRBench-v2~\citep{fu2026ocrbench}, evaluate visual text perception across diverse scenarios, including natural scenes, documents, and multilingual content. However, they provide limited coverage of adversarial OCR patterns.
Existing OCR benchmarks~\citep{mathew2021docvqa,masry2022chartqa,wang2024charxiv}, such as OCRBench~\citep{liu2024ocrbench}, CCOCR~\citep{yang2025cc}, and OCRBench-v2~\citep{fu2026ocrbench}, evaluate visual text perception across diverse scenarios, including natural scenes, documents, and multilingual content. However, they provide limited coverage of adversarial OCR patterns.

% Recent adversarial OCR benchmarks begin to address this gap. AdvOCR~\citep{xu2026vacot} focuses on adversarial OCR robustness but remains limited in scale and taxonomy, while SmuggleBench~\citep{li2026making} evaluates image-level hidden-text extraction without region grounding or designed question answering. In contrast, our proposed AdvSpot provides bounding boxes, transcriptions, perception-type labels, and region-grounded VQA instances under a failure-mechanism-based taxonomy. To the best of our knowledge, it is the first benchmark to evaluate adversarial OCR as a unified \textbf{locate--read--understand} task.

Recent adversarial OCR benchmarks begin to address this gap. AdvOCR~\citep{xu2026vacot} focuses on adversarial OCR robustness but remains limited in scale and taxonomy, while SmuggleBench~\citep{li2026making} evaluates image-level hidden-text extraction without region grounding or designed question answering. 
% In contrast, our proposed AdvSpot provides bounding boxes, transcriptions, perception-type labels, and region-grounded VQA instances under a failure-mechanism-based taxonomy. To the best of our knowledge, it is the first benchmark to evaluate adversarial OCR perception in a grounded setting with region-level annotations and question answering.
In contrast, our proposed AdvSpot provides bounding boxes, transcriptions, region-grounded VQA and perception-type labels instances under a comprehensive failure-mechanism-based taxonomy.

\subsection{Adversarial OCR Perception Methods}

% Existing approaches to adversarial OCR perception mainly recover difficult visual text at inference time. VACoT~\citep{xu2026vacot} adopts a ``thinking-with-images'' paradigm~\citep{chng2025sensenova,zhang2025thyme,zheng2025deepeyes} that dynamically applies visual transformations such as cropping and resizing, while SemVink~\citep{li2025semvink} shows that zoom-out views can reveal text hidden in AI-generated images. Li et al.~\citep{li2026making} further improve hidden-text extraction through detailed chain-of-thought (CoT) prompting. Although effective, these approaches introduce additional inference or reasoning overhead. In contrast, ArmorOCR internalizes transformation-revealed perception during training, enabling single-pass inference on the original image.

Existing approaches to adversarial OCR perception mainly recover difficult visual text at inference time. VACoT~\citep{xu2026vacot} adopts a ``thinking-with-images'' paradigm~\citep{chng2025sensenova,zhang2025thyme,zheng2025deepeyes} that dynamically applies visual transformations such as cropping and resizing, while SemVink~\citep{li2025semvink} shows that zoom-out views can reveal text hidden in AI-generated images. Although effective, these approaches introduce additional inference-time overhead. Li et al.~\citep{li2026making} improve hidden-text extraction through detailed chain-of-thought (CoT) prompting, but such prompt-based strategies require benchmark-specific adaptation. 
In contrast, our ArmorOCR enables single-pass inference on the original image without additional visual transformations or extended reasoning processes~\citep{hu2024visual}.
% In contrast, ArmorOCR internalizes transformation-revealed perception during training, enabling single-pass inference on the original image without additional visual transformations or extended reasoning processes.

% Although effective, these approaches require additional visual operations or reasoning steps during inference. In contrast, ArmorOCR internalizes transformation-revealed perception during training, enabling efficient single-pass inference on the original image without extra test-time transformations or reasoning.

\subsection{Self-Distillation and Reinforcement Learning}

On-policy self-distillation (OPSD)~\citep{zhao2026self} trains the student on its own sampled trajectories, with a teacher equipped with privileged information to provide token-level distribution guidance.
% On-policy self-distillation (OPSD)~\citep{zhao2026self} reduces the token-level distribution gap between the student and a privileged teacher on inference trajectories.
Recent multimodal studies~\citep{yuan2026vision,cai2026thinking,tian2026vicur,kim2025cosmos} further extend privileged information from textual to visual representations. For example, Vision-OPD~\citep{yuan2026vision} transfers fine-grained perception from a crop-conditioned teacher to a full-image student. ArmorOCR extends this principle to adversarial OCR by transferring perception revealed through multiple transformations from a privileged-view teacher to an original-view student.

Reinforcement learning has been widely used for LMM alignment by optimizing task-specific rewards~\citep{schulman2017proximal,guo2025deepseek,zheng2025group,yu2026dapo}, such as GRPO~\citep{guo2025deepseek}, which performs policy optimization using relative rewards among multiple sampled responses without requiring an explicit value model. Such reward-driven optimization is suitable for adversarial OCR perception, where localization, recognition, spotting, and grounded answering can be directly optimized through task-specific rewards.

\begin{table}[h]
\centering

\fontsize{8.5}{9.}\selectfont
\caption{Comparison to prior adversarial OCR benchmarks.}
\label{tab:benchmark_comparison}
\vspace{-6pt}

\setlength{\tabcolsep}{4pt}
\renewcommand{\arraystretch}{1.12}

\begin{tabular}{@{}lccc@{}}
\toprule
\textbf{Dimension}
& \textbf{AdvOCR}
& \textbf{SmuggleBench}
& \textbf{AdvSpot} \\
\midrule

\# Images
& 100
& 1,700
& 383 \\

\# Perception Types
& --
& 6
& \textbf{13} \\

BBox Annotations
& \ding{55}
& \ding{55}
& \ding{51} \\

Region-aware QA
& \ding{55}
& \ding{55}
& \ding{51} \\

\bottomrule
\end{tabular}
% \caption{Comparison with existing adversarial OCR benchmarks.}
\end{table}
\vspace{-13pt}

\section{Benchmark: AdvSpot}
\label{Benchmark: AdvSpot}

% AdvSpot advances adversarial OCR benchmarking in three respects. First, it offers broader and more systematic coverage of adversarial OCR perception types by organizing them according to their underlying failure mechanisms. Second, it introduces fine-grained annotations, including bounding boxes, transcriptions, perception-type labels, and region-grounded question--answer pairs. Third, it evaluates adversarial OCR perception as an integrated locate--read--understand process via region-grounded VQA. Table~\ref{tab:benchmark_comparison} compares AdvSpot with existing adversarial OCR benchmarks.

AdvSpot advances adversarial OCR benchmarking in three respects. First, it offers broader and more systematic coverage of adversarial OCR perception types by organizing them according to their underlying failure mechanisms. Second, it introduces fine-grained annotations, including bounding boxes, transcriptions, perception-type labels, and region-grounded VQA pairs. Third, it provides a region-grounded evaluation framework for adversarial OCR perception, enabling more informative assessment beyond recognition-only evaluation. Table~\ref{tab:benchmark_comparison} compares AdvSpot with existing adversarial OCR benchmarks.

\subsection{Task Definition}

AdvSpot instantiates grounded adversarial OCR perception as a region-grounded VQA task. Each instance is represented as
\begin{equation}
    \mathcal{S}
    =
    (x,\mathcal{C}^\star,b^\star,t^\star,q,a^\star),
\end{equation}
where $x$ is an image containing adversarial OCR patterns, $\mathcal{C}^\star$ is the set of perception-type labels associated with the annotated text region $b^\star$, and $t^\star$ is its transcription. The question $q$ uniquely refers to the region--text pair $(b^\star,t^\star)$ through visual attributes or spatial context, and $a^\star$ is the corresponding answer.

% During evaluation, the model receives only $(x,q)$. Answering correctly therefore requires the model to locate the referenced region, read its adversarial text, and interpret the region-specific query, providing an \textbf{end-to-end} evaluation of locate--read--understand.

During evaluation, the model receives only $(x,q)$. Answering correctly requires the model to ground the question to the relevant text region, recognize the adversarial text, and provide a region-specific answer. This formulation enables grounded evaluation of adversarial OCR perception.

% \begin{figure}[h]
% \centering
% \centerline{\epsfig{figure=Figures/bench_taxonomy.pdf,width=9cm}}
% \caption{Perception types of AdvSpot grouped by their underlying failure mechanisms.}
% \label{fig:advspot_taxonomy_statistics}
% \end{figure}

% % 导言区
% \usepackage{graphicx}

% 正文
% 导言区

% 正文
\begin{figure}[t]
\centering
\begin{subfigure}[t]{0.48\linewidth}
    \centering
    \includegraphics[width=0.92\linewidth]{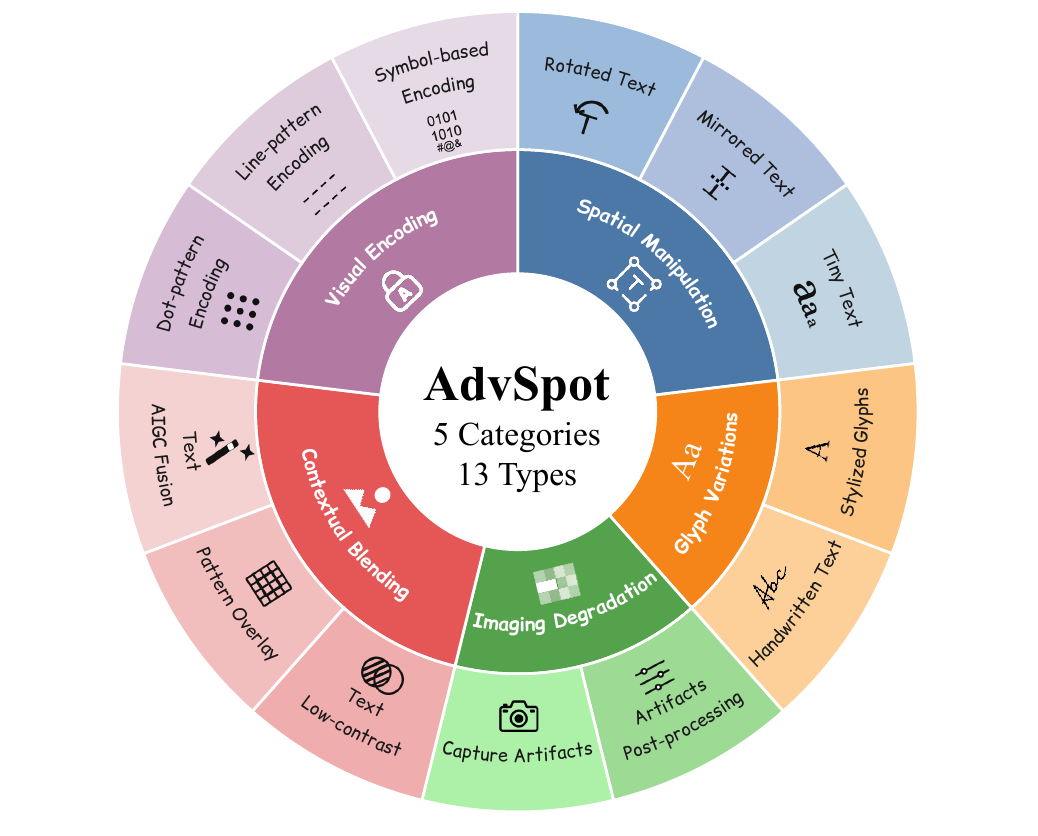}
    \caption{AdvSpot perception types}
    \label{fig:advspot_taxonomy}
\end{subfigure}
\hfill
\begin{subfigure}[t]{0.48\linewidth}
    \centering
    \includegraphics[width=\linewidth]{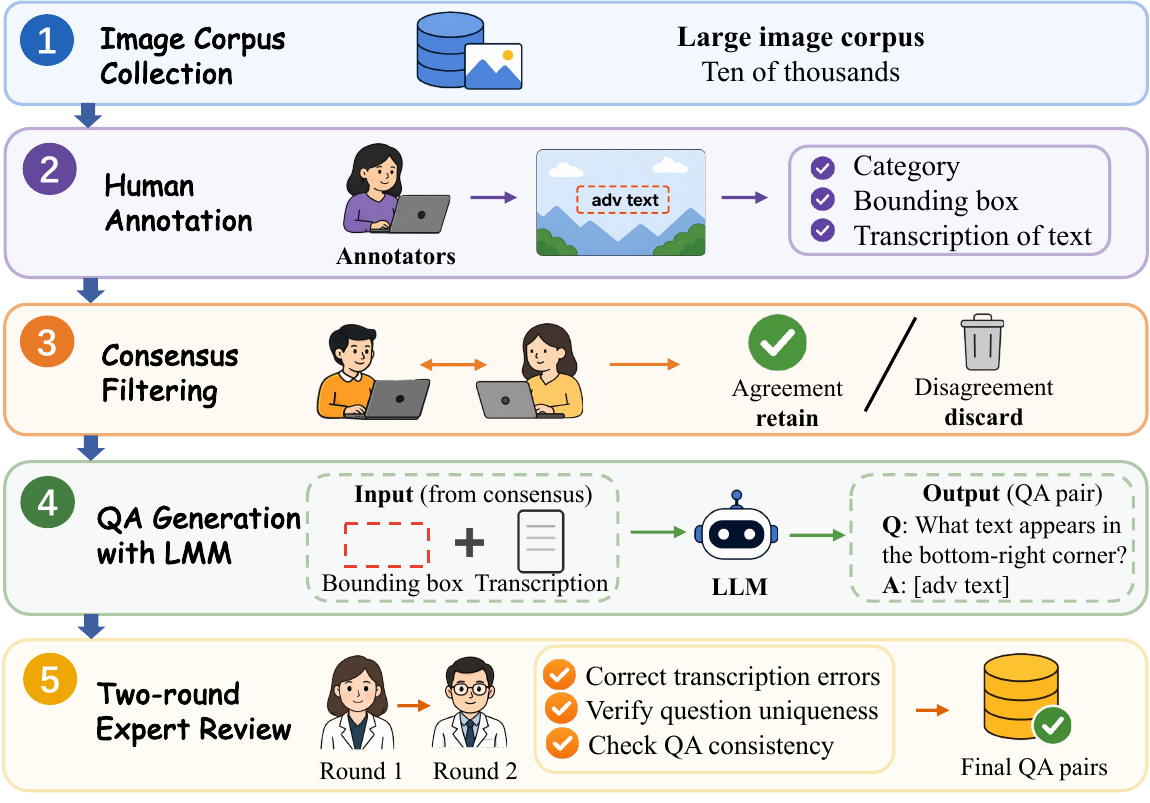}
    \caption{Human-in-the-loop pipeline}
    \label{fig:advspot_pipeline}
\end{subfigure}
\caption{Overview of AdvSpot: perception types grouped by underlying
failure mechanisms and the human-in-the-loop construction pipeline.}
\label{fig:advspot_overview}
\end{figure}

\vspace{-10pt}

\subsection{Taxonomy}

% AdvSpot organizes adversarial OCR patterns according to the underlying mechanisms that cause perception failures, providing a systematic basis for pattern coverage and category-wise diagnosis. It covers 5 primary categories and 13 fine-grained adversarial OCR types, as summarized in Figure~\ref{fig:advspot_taxonomy_statistics}, with detailed definitions provided in Appendix~A.1. Specifically, \textbf{Spatial Manipulation} introduces geometric variations that challenge text orientation and scale perception; \textbf{Glyph Variation} alters character appearances beyond standard glyph patterns; \textbf{Image Degradation} reduces visual quality through capture and post-processing artifacts; \textbf{Contextual Blending} integrates text with surrounding visual contexts, making it difficult to distinguish from the background; and \textbf{Visual Encoding} represents text through non-standard visual patterns such as symbols, dots, and lines rather than conventional glyph shapes.
% AdvSpot comprises 5 primary categories and \textbf{13 fine-grained} adversarial OCR types grouped by perception-failure mechanisms, as summarized in Figure~\ref{fig:advspot_taxonomy_statistics}, with detailed definitions and examples provided in Appendix~A.1.

AdvSpot comprises 5 primary categories and \textbf{13 fine-grained} adversarial OCR types grouped by perception-failure mechanisms, as summarized in Figure~\ref{fig:advspot_taxonomy}. The number of subtypes in each category, along with detailed definitions and examples, is provided in Appendix~\ref{Detailed Definition of each Taxonomy}.

\subsection{Annotation and QA Construction}
\label{Annotation and QA Construction}

We construct AdvSpot through a \textit{human-in-the-loop} pipeline, as illustrated in Figure~\ref{fig:advspot_pipeline}. We first collect tens of thousands of \textcolor{black}{real-world} candidate images potentially containing adversarial OCR patterns. Each image is independently annotated by two annotators with bounding boxes, transcriptions, and perception-type labels according to our taxonomy. We retain only regions with consistent annotations from both annotators and discard images without agreed regions. For each retained region, we use Qwen3-VL-235B-A22B-Instruct to generate a region-grounded question--answer pair conditioned on the image, bounding box, and transcription; the detailed prompt is provided in Appendix~\ref{Detailed Construction of VQA Pairs}. Each QA pair then undergoes two rounds of expert review to correct errors, verify that the question uniquely refers to the target region, and ensure that the answer is consistent with the verified transcription.

% \begin{figure}[h]
% \centering
% \centerline{\epsfig{figure=Figures/benchmark_pipeline.pdf,width=9cm}}
% % \caption{Human-in-the-loop construction pipeline of our AdvSpot. The process includes image corpus collection, adversarial OCR annotation, consensus filtering, grounded QA generation, and two-round human verification.}
% \caption{Human-in-the-loop pipeline of our AdvSpot.}
% \label{fig:pipeline}
% \end{figure}

\subsection{Evaluation Protocol}

We report VQA accuracy as the primary metric and the intersection over union (IoU) between predicted and ground-truth bounding boxes as a localization metric:
\begin{equation}
\label{eq:eval_metrics}
\mathrm{Acc}(\hat a,a^\star)
=\mathbb{I}\!\left[a^\star\preceq_{\mathrm{sub}}\hat a\right],
\qquad
\mathrm{IoU}(\hat b,b^\star)
=\frac{|\hat b\cap b^\star|}{|\hat b\cup b^\star|}.
\end{equation}
Here, $u\preceq_{\mathrm{sub}}v$ means that $u$ is a contiguous
substring of $v$; thus, additional text is allowed in the prediction.
\textcolor{black}{
We additionally report Joint@0.5, the percentage of instances
satisfying both $\mathrm{Acc}(\hat a,a^\star)=1$ and
$\mathrm{IoU}(\hat b,b^\star)\geq 0.5$,
to evaluate joint answer and localization correctness.
}
Evaluation prompts are provided in Appendix~\ref{Detailed Testing Prompt}.

\section{Method: ArmorOCR}
\label{Method: ArmorOCR}
% We propose \textbf{ArmorOCR} (Figure ~\ref{fig:ArmorOCR_framework}), a robust framework for adversarial text perception via observation transfer and spotting refinement. ArmorOCR consists of two stages. Stage~1 transfers robust perception from privileged transformed observations to the student through On-Policy Self-Distillation. Stage~2 further refines the model with task-conditioned spotting rewards that strengthen localization, recognition, full spotting, and grounded VQA. 

% We propose \textbf{ArmorOCR} (Figure~\ref{fig:ArmorOCR_framework}), a two-stage framework for adversarial OCR perception via observation transfer and reward-driven refinement. ArmorOCR is based on the observation that adversarial OCR failures arise from two challenges: (1) the original view may hide critical visual cues required to perceive adversarial text, and (2) adversarial perception alone is insufficient for completing grounded OCR tasks that require localization, recognition, and understanding. Therefore, ArmorOCR first transfers transformation-revealed perception from privileged observations to the student through OPSD, and then optimizes the model toward the locate--read--understand objective through GRPO with task-conditioned rewards.

\begin{figure*}[t]
\centering
\centerline{\epsfig{figure=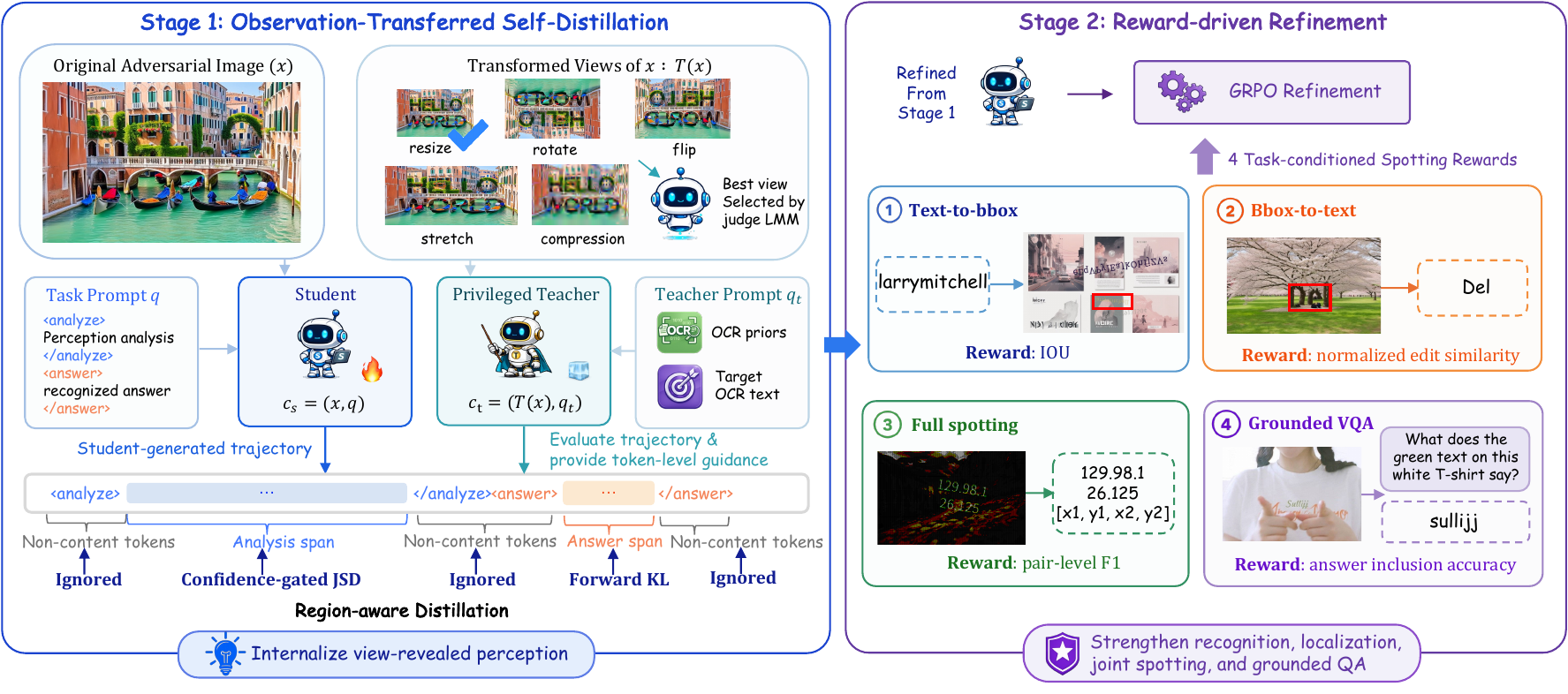,width=14cm}}
% \caption{Overview of \textbf{ArmorOCR}, a two-stage framework for grounded adversarial OCR perception. Stage~1 transfers transformation-induced perception from privileged observations via OPSD, while Stage~2 optimizes locate--read--understand capabilities with task-conditioned GRPO rewards. Except for resizing, all transformations are applied at the original image resolution; reduced sizes are only for visualization.}
\caption{Overview of \textbf{ArmorOCR}, a two-stage framework for grounded adversarial OCR perception. Stage~1 transfers transformation-induced perception from privileged observations via OPSD, while Stage~2 optimizes grounded OCR perception through task-conditioned GRPO rewards for localization, recognition, full spotting, and grounded VQA.}
% Except for resizing, all transformations are applied at the original image resolution; reduced sizes are only for visualization.
\label{fig:ArmorOCR_framework}
\end{figure*}

We propose \textbf{ArmorOCR} (Figure~\ref{fig:ArmorOCR_framework}), a two-stage framework for adversarial OCR perception via observation transfer and reward-driven refinement. ArmorOCR is based on the observation that adversarial OCR failures arise from two challenges: (1) the original view may hide critical visual cues required to perceive adversarial text, and (2) adversarial perception alone is insufficient for completing grounded OCR tasks that require localization, recognition, and grounded answering. Therefore, ArmorOCR first transfers transformation-revealed perception from privileged observations to the student through OPSD, and then optimizes grounded OCR perception through GRPO with task-conditioned rewards.

% Adversarial OCR perception requires two key abilities: perceiving text signals that may become salient only under certain image transformations, and accurately localizing adversarial text regions. Existing paradigms, such as thinking with images~\citep{}, recover such perception by repeatedly invoking visual tools, which inevitably introduces additional latency. In contrast, ArmorOCR aims to internalize both transformation-induced visual perception and text localization ability into the model during training, enabling efficient single-pass inference. ArmorOCR consists of two stages. Stage~1 transfers robust perception from privileged transformed observations to the student through On-Policy Self-Distillation (OPSD). Stage~2 further refines the model with task-conditioned spotting rewards that strengthen recognition, localization, full spot-text prediction, and grounded VQA answering.

% \section{Method: ArmorOCR}

% We propose \textbf{ArmorOCR} (Figure~\ref{fig:ArmorOCR_framework}), a two-stage framework that combines transformation-revealed perception internalization with grounded capability refinement. Stage~1 addresses the rollout cold-start problem by transferring privileged transformed-view perception to the student through OPSD. Stage~2 moves beyond teacher imitation and directly optimizes localization, recognition, full spotting, and grounded VQA through GRPO with task-conditioned rewards.

% \vspace{-10pt}

\subsection{Stage~1: Observation-Transferred Self-Distillation}
% This design is motivated by the observation that adversarial OCR difficulty is often view-dependent. For example, in AIGC synthesis attacks, the original image may cause the model to focus on high-frequency scenery while ignoring global text structures. Downscaling can suppress irrelevant high-frequency details and make the hidden text more salient. ArmorOCR transfers such transformation-induced perception advantages into the student.

% Motivated by the observation that adversarial OCR difficulty is often view-dependent~\citep{li2025semvink}, we propose an Observation-Transferred Self-Distillation (OTSD) paradigm. 

Motivated by the observation that adversarial OCR difficulty is often view-dependent~\citep{li2025semvink}, we aim to transfer the perceptual advantages revealed by transformed views into the model during training. Building upon OPSD, we propose Observation-Transferred Self-Distillation (OTSD), where a teacher conditioned on privileged transformed views guides the student.

% Let $x$ be the original adversarial image and $q$ be the task prompt. The prompt $q$ instructs the model to provide an intermediate perception analysis within \texttt{<analyze>} and \texttt{</analyze>} tags, followed by the final recognized answer within \texttt{<answer>} and \texttt{</answer>} tags. The student is conditioned only on the original image and the task prompt:
% \begin{equation}
%     c_s = (x, q).
% \end{equation}
% In contrast, the teacher is conditioned on privileged observations:
% \begin{equation}
%     c_t = (\mathcal{T}(x), q_t),
% \end{equation}
% where $\mathcal{T}(x)$ denotes a transformed view of $x$. We consider five transformations: resizing, stretching, rotation, flipping, and compression. For each image, the judge LMM, Qwen3-VL-235B-A22B-Instruct, selects the transformed view with the highest recognition accuracy. 

\begin{table*}[t]
\centering
\fontsize{6.8pt}{7.6pt}\selectfont
\caption{Fine-grained accuracy and average localization IoU on AdvSpot. All values are reported in \%. The best and second-best results are highlighted in \textbf{bold} and \underline{underlined}, respectively. \textbf{Avg. Acc.} and \textbf{Avg. IoU} denote the sample-weighted averages of accuracy and IoU, respectively. Category-wise IoU results are provided in \textbf{Appendix~\ref{More Experiments on AdvSpot}}. The same conventions apply to subsequent tables.}
\label{tab:advspot-results}
\vspace{-6pt}
\renewcommand{\arraystretch}{0.96}
\setlength{\tabcolsep}{2.4pt}

\begin{tabularx}{\textwidth}{@{}ll*{9}{>{\centering\arraybackslash}X}@{}}
    \toprule
    \multirow{2}{*}{\textbf{Category}}
    & \multirow{2}{*}{\textbf{Subtype}}
    & \multicolumn{4}{c}{\textit{\textbf{Open-source Qwen3-VL}}}
    & \multicolumn{4}{c}{\textit{\textbf{Proprietary LMMs}}}
    & \multicolumn{1}{c}{\textit{\textbf{Our LMM}}} \\
    \cmidrule(lr){3-6}\cmidrule(lr){7-10}\cmidrule(l){11-11}
    &
    & \multicolumn{1}{c}{\makecell{8B\\\textit{(base)}}}
    & \multicolumn{1}{c}{\makecell{30B-\\A3B}}
    & \multicolumn{1}{c}{32B}
    & \multicolumn{1}{c}{\makecell{235B-\\A22B}}
    & \multicolumn{1}{c}{GPT-5.1}
    & \multicolumn{1}{c}{\makecell{Gemini\\3 Flash}}
    & \multicolumn{1}{c}{\makecell{GLM\\5.3 Flash}}
    & \multicolumn{1}{c}{Kimi-K3}
    & \multicolumn{1}{c}{\makecell{\textbf{Armor-}\\\textbf{OCR}}} \\
    \midrule

    \multirow{2}{*}{\makecell[l]{Imaging\\Degradation}}
    & Capture Artifacts
    & 40.0 & 53.3 & 53.3 & \underline{63.3} & 26.7 & \textbf{66.7} & 50.0 & 53.3 & 43.3 \\
    & Post-processing
    & 53.3 & 53.3 & 50.0 & \underline{60.0} & 43.3 & \textbf{80.0} & 40.0 & 56.7 & \underline{60.0} \\
    \midrule

    \multirow{3}{*}{\makecell[l]{Spatial\\Manipulation}}
    & Rotated Text
    & 48.6 & 48.6 & 54.3 & \underline{60.0} & 28.6 & \textbf{62.9} & 48.6 & 51.4 & 54.3 \\
    & Mirrored Text
    & 20.0 & 26.7 & 20.0 & 23.3 & 13.3 & 13.3 & \underline{43.3} & 30.0 & \textbf{46.7} \\
    & Tiny Text
    & \underline{60.0} & 54.3 & 51.4 & 57.1 & 45.7 & \textbf{68.6} & 45.7 & \textbf{68.6} & 57.1 \\
    \midrule

    \multirow{2}{*}{\makecell[l]{Glyph\\Variations}}
    & Stylized Glyphs
    & 30.0 & 50.0 & 46.7 & 53.3 & 30.0 & 56.7 & \underline{60.0} & \textbf{70.0} & 26.7 \\
    & Handwritten Text
    & 36.7 & 53.3 & 40.0 & 53.3 & 30.0 & \underline{56.7} & 53.3 & \textbf{60.0} & 46.7 \\
    \midrule

    \multirow{3}{*}{\makecell[l]{Visual\\Encoding}}
    & Symbol Encoding
    & 0.0 & 0.0 & 0.0 & 5.7 & \underline{31.4} & 20.0 & 5.7 & 2.9 & \textbf{68.6} \\
    & Dot Encoding
    & 23.3 & 23.3 & 20.0 & 23.3 & \underline{33.3} & \underline{33.3} & 26.7 & 13.3 & \textbf{63.3} \\
    & Line Encoding
    & 2.9 & 5.7 & 5.7 & \underline{14.3} & 11.4 & 11.4 & 0.0 & 2.9 & \textbf{54.3} \\
    \midrule

    \multirow{3}{*}{\makecell[l]{Contextual\\Blending}}
    & AIGC Fusion Text
    & 5.7 & 8.6 & 5.7 & 11.4 & 5.7 & \underline{17.1} & 11.4 & 5.7 & \textbf{82.9} \\
    & Low Contrast Text
    & 45.7 & 45.7 & \underline{48.6} & \underline{48.6} & 28.6 & \textbf{74.3} & 45.7 & \underline{48.6} & 42.9 \\
    & Pattern Overlay
    & 26.7 & 30.0 & 23.3 & 26.7 & 13.3 & \underline{36.7} & \underline{36.7} & 3.3 & \textbf{66.7} \\
    \midrule

    \rowcolor{gray!10}[8pt][8pt]
    Avg. \textbf{Acc.}
    & --
    & 29.8 & 34.7 & 31.9 & 38.3 & 26.5 & \underline{45.2} & 35.2 & 35.0 & \textbf{57.1} \\
    \midrule

    \rowcolor{gray!10}[8pt][8pt]
    Avg. \textbf{IoU}
    & --
    & 54.4 & 55.0 & \underline{61.1} & 58.6 & 27.8 & 59.2 & 40.6 & 54.9 & \textbf{63.2} \\

    \midrule
    \rowcolor{gray!10}[8pt][8pt]
    \textcolor{black}{\textbf{Joint@0.5}}
    & --
    & \textcolor{black}{24.7}
    & \textcolor{black}{28.8}
    & \textcolor{black}{28.8}
    & \textcolor{black}{28.0}
    & \textcolor{black}{6.7}
    & \textcolor{black}{\underline{29.0}} 
    & \textcolor{black}{15.9}
    & \textcolor{black}{23.7}
    & \textcolor{black}{\textbf{46.0}} \\

    \bottomrule
\end{tabularx}

\end{table*}

Let $x$ be the original adversarial image and $q$ be the task prompt.
The prompt $q$ instructs the model to provide an intermediate perception
analysis within \texttt{<analyze>} and \texttt{</analyze>} tags, followed by
the final recognized answer within \texttt{<answer>} and
\texttt{</answer>} tags. The student sees the original image, whereas
the teacher receives a privileged transformed view:
\begin{equation}
    c_s=(x,q), \qquad c_t=(\mathcal{T}(x),q_t).
\end{equation}
Here, $\mathcal{T}(x)$ denotes a transformed view of $x$. We consider
five transformations: resizing, stretching, rotation, flipping, and
compression. For each image, the judge LMM,
Qwen3-VL-235B-A22B-Instruct, selects the transformed view with the
highest recognition accuracy.

The teacher prompt $q_t$ further extends $q$ with adversarial OCR priors and the target OCR text, enabling the teacher to understand the adversarial visual pattern and provide reliable token-level answer guidance. Further details are provided in Appendix~\ref{More Details of Stage 1}.
% The teacher prompt $q_t$ augments $q$ with the ground-truth transcription $t^\star$, to provide reliable token-level answer guidance. Further details are provided in Appendix~B.1.

The student first samples an on-policy response:
\begin{equation}
    \hat{y} \sim \pi_{\theta}(\cdot \mid c_s).
\end{equation}
The teacher then evaluates the student-generated trajectory under privileged observations. At token position $i$, the student and teacher distributions are:
\begin{equation}
    p_s^i = \pi_{\theta}(\cdot \mid c_s, \hat{y}_{<i}), \quad
    p_t^i = \pi_{\theta}'(\cdot \mid c_t, \hat{y}_{<i}).
\end{equation}

Vanilla OPSD uniformly distills all response tokens, overlooking their distinct roles in the final OCR prediction. Analysis tokens encode potentially uncertain intermediate perception, answer tokens directly determine the transcription, and structural tokens contain no task-relevant content. We therefore introduce a \textbf{response-region-aware distillation loss}, partitioning the response into an analysis span $\mathcal{R}_{\text{ana}}$, an answer span $\mathcal{R}_{\text{ans}}$, and non-content structural tokens using predefined tags.

For analysis tokens, uniformly applying distillation may propagate uncertain or noisy intermediate reasoning. We therefore use confidence-gated Jensen--Shannon divergence (JSD)~\citep{lu2026self} to selectively transfer reliable teacher signals:
\begin{equation}
    \mathcal{L}_{\text{ana}}^{i}
    =
    g_i D_{\mathrm{JSD}}^{\beta}(p_t^i \Vert p_s^i),
    \quad i \in \mathcal{R}_{\text{ana}},
\end{equation}
where $D_{\mathrm{JSD}}^{\beta}$ denotes the generalized JSD with mixture coefficient $\beta$. The token-level confidence weight is defined as
\begin{equation}
    g_i =
    \sigma \left(
    \gamma
    \left[
    \max_k \log p_t^i(k)
    -
    \max_k \log p_s^i(k)
    \right]
    \right),
\end{equation}
where $\sigma(\cdot)$ is the sigmoid function and $\gamma$ controls the sharpness of the confidence gate. The weight $g_i$ increases when the privileged teacher is more confident than the student, strengthening distillation at positions with reliable teacher guidance while suppressing uncertain signals.

For answer tokens, which directly determine the final OCR transcription, we apply stronger supervision using forward KL divergence:
\begin{equation}
    \mathcal{L}_{\text{ans}}^{i}
    =
    \mathrm{KL}(p_t^i \Vert p_s^i),
    \quad i \in \mathcal{R}_{\text{ans}}.
\end{equation}
Conditioned on privileged observations and OCR guidance, the teacher provides a reliable target answer distribution. We assign uniform weights to all answer tokens, as each contributes to the final transcription.

Non-content tokens, such as structural markers, are excluded from the loss because they encode only the output format and do not contribute to adversarial text perception. The resulting objective selectively transfers intermediate perception while imposing direct supervision on the final OCR output:
\begin{equation}
\mathcal{L}_{\mathrm{OTSD}}
=
\frac{
\sum_{i \in \mathcal{R}_{\text{ana}}}
\mathcal{L}_{\text{ana}}^{i}
+
\sum_{i \in \mathcal{R}_{\text{ans}}}
\mathcal{L}_{\text{ans}}^{i}
}{
\sum_{i \in \mathcal{R}_{\text{ana}}} g_i
+
|\mathcal{R}_{\text{ans}}|
+
\epsilon
}.
\end{equation}

\begin{table*}[t]
    \centering
    \fontsize{7.2pt}{8.5pt}\selectfont
    \caption{Results on other adversarial OCR benchmarks: AdvOCR and SmuggleBench.}
    \label{tab:other_adv}
    \vspace{-6pt}
    \renewcommand{\arraystretch}{0.96}
    \setlength{\tabcolsep}{2.4pt}
    \begin{tabularx}{\textwidth}{@{}ll*{9}{>{\centering\arraybackslash}X}@{}}
        \toprule
        \multirow{2}{*}{\textbf{Benchmark}}
        & \multirow{2}{*}{\textbf{Task}}
        & \multicolumn{4}{c}{\textit{\textbf{Open-source Qwen3-VL}}}
        & \multicolumn{2}{c}{\textit{\textbf{Closed-source LMMs}}}
        & \multicolumn{3}{c}{\textit{\textbf{Specialized LMMs}}} \\
        \cmidrule(lr){3-6}\cmidrule(lr){7-8}\cmidrule(l){9-11}
        &
        & \multicolumn{1}{c}{\makecell{8B\\{\fontsize{6.2pt}{7pt}\selectfont\textit{(Our Base)}}}}
        & \multicolumn{1}{c}{\makecell{30B-\\A3B}}
        & \multicolumn{1}{c}{32B}
        & \multicolumn{1}{c}{\makecell{235B-\\A22B}}
        & \multicolumn{1}{c}{GPT-5}
        & \multicolumn{1}{c}{\makecell{Gemini\\2.5 Flash}}
        & \multicolumn{1}{c}{VACoT}
        & \multicolumn{1}{c}{\makecell{Smuggle-\\CoT}}
        & \multicolumn{1}{c}{\makecell{\textbf{Armor-}\\\textbf{OCR}}} \\
        \midrule

        \multirow{3}{*}{AdvOCR}
        & Real-world
        & 30.0 & 34.0 & \underline{44.0} & \underline{44.0}
        & 12.0 & 34.0 & \textbf{62.0} & -- & \underline{44.0} \\

        & Synthetic
        & 12.0 & 12.0 & 8.0 & 12.0
        & 22.0 & 12.0 & \underline{48.0} & -- & \textbf{68.0} \\

        \rowcolor{gray!10}[8pt][8pt]
        \cellcolor{white} & Avg.
        & 21.0 & 23.0 & 26.0 & 28.0
        & 17.0 & 23.0 & \underline{55.0} & -- & \textbf{56.0} \\

        \midrule

        \multirow{7}{*}{SmuggleBench}
        & Tiny Text
        & 26.1 & 25.9 & 23.6 & 25.1
        & 18.2 & 26.3 & -- & \underline{30.2} & \textbf{33.0} \\

        & Occluded Text
        & 18.2 & 23.2 & 19.1 & 19.6
        & 9.7 & \underline{24.1} & -- & \textbf{26.1} & 17.5 \\

        & Low Contrast
        & 3.0 & \underline{8.8} & 3.5 & 6.5
        & 4.0 & \textbf{12.5} & -- & 8.5 & 6.5 \\

        & Handwritten
        & \underline{31.7} & 30.4 & 29.3 & \textbf{33.7}
        & 15.8 & 15.4 & -- & \underline{31.7} & 27.0 \\

        & Artistic
        & 14.1 & 14.5 & 11.6 & 15.6
        & 11.1 & 8.7 & -- & \underline{18.6} & \textbf{19.0} \\

        & AI Illusions
        & 0.0 & 0.3 & \underline{0.8} & \underline{0.8}
        & 0.3 & 0.0 & -- & 0.0 & \textbf{8.2} \\

        \rowcolor{gray!10}[8pt][8pt]
        \cellcolor{white} & Avg.
        & 13.3 & 14.8 & 14.3 & 14.6
        & 8.5 & 12.4 & -- & \underline{16.4} & \textbf{17.1} \\
        \bottomrule
    \end{tabularx}
\end{table*}
% \vspace{-20pt}

% \subsection{Stage~2: Reward-driven Refinement}

% With the acquired adversarial OCR perception, we further optimize the model toward the locate--read--understand objective through GRPO. Specifically, task-conditioned rewards are designed to explicitly optimize localization, recognition, full spotting, and grounded VQA capabilities.

% Specifically, we construct four types of training tasks: text-to-bbox, bbox-to-text, full-spotting, and VQA, each associated with a task-specific reward. These tasks enhance adversarial OCR perception through localization, recognition, joint spotting, and grounded question answering.

\subsection{Stage~2: Reward-driven Refinement}

% Stage~1 establishes the model's ability to perceive adversarial text, but teacher-guided distillation remains constrained by teacher supervision and does not directly optimize the structured capabilities required by grounded OCR tasks. We therefore apply GRPO with task-conditioned rewards to refine the model toward the locate--read--understand objective. Specifically, we construct four complementary tasks: text-to-bbox for localization, bbox-to-text for recognition, full spotting for joint region--text prediction, and grounded VQA for understanding.

Stage~1 establishes the model's ability to perceive adversarial text, but teacher-guided distillation remains constrained by teacher supervision and does not directly optimize the diverse objectives required by grounded OCR tasks. We therefore apply GRPO with task-conditioned rewards to align the model with grounded adversarial OCR perception. Specifically, we construct four complementary tasks: text-to-bbox for localization, bbox-to-text for recognition, full spotting for joint region--text prediction, and grounded VQA for region-specific question answering.

For \textbf{text-to-bbox}, the model performs text-conditioned region localization by predicting the bounding box corresponding to a given text string. The localization reward is defined as the IoU between the predicted and ground-truth bounding boxes:
\begin{equation}
R_{\mathrm{t2b}}
=
\mathrm{IoU}(\hat{b}, b^\star).
\end{equation}

% where $\hat{b}$ and $b^\star$ denote the predicted and ground-truth bounding boxes, respectively.

For \textbf{bbox-to-text}, the model performs region-conditioned text recognition by predicting the transcription corresponding to a given bounding box. Let $\hat{t}$ and $t^\star$ denote the predicted and ground-truth transcriptions, respectively. We adopt normalized Levenshtein similarity as the recognition reward:
\begin{equation}
R_{\mathrm{b2t}}
=
1 -
\frac{
\mathrm{ED}(\hat{t}, t^\star)
}{
\max(|\hat{t}|, |t^\star|)
},
\end{equation}
where $\mathrm{ED}(\cdot,\cdot)$ denotes the Levenshtein edit distance.

% For \textbf{full spotting}, the model jointly predicts multiple text regions and their corresponding transcriptions. A predicted bbox-text pair is considered a valid match only when both localization and recognition constraints are satisfied:
% \begin{equation}
% \mathrm{IoU}(\hat{b}, b^\star) \geq \theta_{\mathrm{iou}},
% \quad
% \mathrm{Sim}(\hat{t}, t^\star) \geq 1-\theta_{\mathrm{edit}},
% \label{eq_1}
% \end{equation}
% where $\mathrm{Sim}(\cdot,\cdot)$ denotes the normalized Levenshtein similarity, and $\theta_{\mathrm{iou}}$ and $\theta_{\mathrm{edit}}$ are predefined thresholds for localization and recognition matching, respectively. We perform greedy matching between predicted and ground-truth pairs and use pair-level F1 score as the joint spotting reward:
% \begin{equation}
% R_{\mathrm{joint}}
% =
% \frac{2PR}{P+R},
% \end{equation}
% where $P$ and $R$ denote the precision and recall of matched bbox-text pairs.

For \textbf{full spotting}, the model jointly predicts text regions and their
corresponding transcriptions. We greedily match predicted and reference pairs
if both the localization and recognition criteria are satisfied, and use
pair-level F1 as the reward:
\begin{equation}
\label{eq:joint_reward}
\mathrm{IoU}(\hat b,b^\star)\geq\theta_{\mathrm{iou}},
\quad
\mathrm{Sim}(\hat t,t^\star)\geq 1-\theta_{\mathrm{edit}};
\qquad
R_{\mathrm{joint}}=\frac{2PR}{P+R}.
\end{equation}
We set $\theta_{\mathrm{iou}}=0.5$ and $\theta_{\mathrm{edit}}=0.1$;
$\mathrm{Sim}(\cdot,\cdot)$ denotes normalized Levenshtein similarity,
and $P$ and $R$ denote precision and recall over matched pairs.

For \textbf{grounded VQA}, the model answers questions grounded on the annotated adversarial text region. We use normalized answer inclusion as the reward:
\begin{equation}
R_{\mathrm{vqa}}
=
\mathbb{I}
[
a^\star
\preceq_{\mathrm{sub}}
\hat{a}
].
\end{equation}

% The final reward is selected according to the task type $\tau$:
Given the sampled response $\hat{y}$, its ground-truth target $y^\star$, and the task type $\tau$, the final reward is defined as:
\begin{equation}
R(\hat{y},y^\star,\tau)
=
\begin{cases}
R_{\mathrm{t2b}}, & \tau=\texttt{text\_to\_bbox},\\
R_{\mathrm{b2t}}, & \tau=\texttt{bbox\_to\_text},\\
R_{\mathrm{joint}}, & \tau=\texttt{full\_spotting},\\
R_{\mathrm{vqa}}, & \tau=\texttt{vqa}.
\end{cases}
\end{equation}

% Through these task-conditioned rewards, the refinement stage aligns training with the core capabilities required by adversarial OCR perception: localization, recognition, joint spot-text grounding, and grounded VQA understanding.

\begin{figure}[h]
\centering

\begin{minipage}[t]{0.54\linewidth}
\vspace{0pt}
\centering
\includegraphics[width=\linewidth]{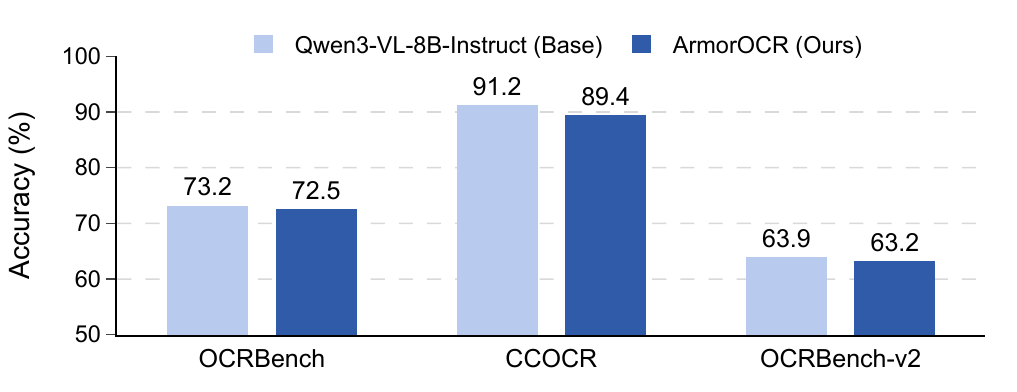}
\vspace{-15pt}
\caption{General OCR performance of ArmorOCR.}
\label{fig:general_ocr_capability}
\end{minipage}
\hfill
\begin{minipage}[t]{0.45\linewidth}
\vspace{0pt}
\centering
\captionof{table}{Comparison of Stage~1 training methods on AdvSpot.}
\label{tab:stage1_methods}
\vspace{-3pt}
\scriptsize
\renewcommand{\arraystretch}{1.12}
\begin{tabular*}{0.85\linewidth}{
    @{\extracolsep{\fill}}lcc@{\hspace{5pt}}
}
\toprule
\textbf{Method} & \textbf{IoU} & \textbf{Acc.} \\
\midrule
Base model          & 54.4 & 29.8 \\
SFT                 & 49.2 & 32.9 \\
OPSD                & 54.5 & 33.9 \\
GRPO                & 56.6 & 37.3 \\
\textbf{OTSD (ours)} & \textbf{58.7} & \textbf{46.8} \\
\bottomrule
\end{tabular*}
\end{minipage}

\end{figure}

\vspace{-10pt}
\section{Experiments}

\subsection{Experimental Setup}

\noindent\textbf{Benchmarks.}
We evaluate models on two groups of benchmarks. For \textbf{adversarial OCR perception}, we evaluate models on our proposed AdvSpot and two existing adversarial OCR-related benchmarks, AdvOCR~\citep{xu2026vacot} and SmuggleBench~\citep{li2026making}. For \textbf{general OCR perception}, we further evaluate on three widely used general OCR benchmarks: CCOCR~\citep{yang2025cc}, OCRBench~\citep{liu2024ocrbench}, and OCRBench-v2~\citep{fu2026ocrbench}.

% \noindent\textbf{Models.}
% We compare open-source and proprietary LMMs. For open-source models, we evaluate the Qwen3-VL series at different scales~\citep{bai2025qwen3}. For proprietary models, we include Claude-Sonnet-4.5~\citep{claude-4.5}, GPT-4o~\citep{achiam2023gpt}, GPT-5~\citep{gpt5} and Gemini-2.5 Flash~\citep{comanici2025gemini}. Additional benchmark results are reported in Appendix~\ref{More Experiments on AdvSpot}. For existing adversarial OCR benchmarks, we also compare benchmark-specific methods: VACoT~\citep{xu2026vacot} for AdvOCR and Smuggle-CoT~\citep{li2026making} for SmuggleBench\footnote{Since VACoT is not open-sourced, we report its official results. Smuggle-CoT refers to the SmuggleBench-specific CoT-based test-time scaling method using Qwen3-VL-235B-A22B-Instruct with detailed prompting.}. 
\noindent\textbf{Models.}
We compare open-source and proprietary LMMs. For open-source models, we evaluate the Qwen3-VL series at different scales~\citep{bai2025qwen3}. For proprietary models, we include GPT-5.1~\citep{gpt5.1} and Gemini-3 Flash~\citep{comanici2025gemini}, GLM-5.3-Flash~\citep{zeng2026glm} and Kimi-K3~\citep{team2026kimi}. Additional benchmark results are reported in Appendix~\ref{More Experiments on AdvSpot}. For existing adversarial OCR benchmarks, we also compare benchmark-specific methods: VACoT~\citep{xu2026vacot} for AdvOCR and Smuggle-CoT~\citep{li2026making} for SmuggleBench\footnote{Since VACoT is not open-sourced, we report its official results. Smuggle-CoT refers to the SmuggleBench-specific CoT-based test-time scaling method using Qwen3-VL-235B-A22B-Instruct with detailed prompting.}.

\noindent\textbf{Training Details.} Due to the scarcity of large-scale adversarial OCR data, we synthesize stage-specific training sets with controllable adversarial patterns. Stage~1 uses 50K examples to establish adversarial OCR perception, while Stage~2 uses 70K samples enriched with harder adversarial cases for fine-grained grounded OCR optimization. ArmorOCR is built on Qwen3-VL-8B-Instruct. In Stage~1, we keep the teacher frozen and train the student with OTSD. In Stage~2, we apply GRPO with task-conditioned rewards and sample eight responses per prompt. Both stages are trained on PPU-810E accelerators. \textcolor{black}{All benchmarks are evaluated zero-shot.} Data generation and training configurations are detailed in Appendix~\ref{Training Data Construction} and~\ref{Training Configurations}, respectively.

% The IoU threshold $\theta_{\mathrm{iou}}$ and the edit-distance threshold $\theta_{\mathrm{edit}}$ are set to 0.5 and 0.1, respectively. For both stages, the model is trained for three epochs on 16 PPU-810E accelerators.

% Due to the scarcity of large-scale adversarial OCR data, we construct automated data synthesis pipelines to generate stage-specific training sets with controllable adversarial patterns. Stage~1 uses 50K samples for observation-transferred self-distillation, while Stage~2 uses 70K samples with more challenging adversarial patterns and grounded spotting objectives. Detailed data generation procedures are provided in Appendix C.1. 

\noindent\textbf{Metrics.}
For AdvSpot, we report accuracy and IoU\textcolor{black}{, together with Joint@0.5}. For AdvOCR, we report pass@1 accuracy evaluated by Qwen3-30B-A3B-2507~\citep{bai2025qwen3}, following its official evaluation protocol. For SmuggleBench, we focus on its ``Perceptual Blindness'' subset, which is most aligned with our task, and report the TER metric introduced in the original paper. For the three general OCR benchmarks, we adopt the open-source evaluation toolkit VLMEvalKit~\citep{duan2024vlmevalkit}, whose results are largely consistent with the official metrics, and report the overall benchmarks scores.

\subsection{Performance Analysis}

\noindent\textbf{Results on AdvSpot.}
As shown in Table~\ref{tab:advspot-results}, both open-source and proprietary LMMs struggle with several AdvSpot subtypes. Symbol Encoding and AIGC Fusion Text remain highly challenging, with most models achieving nearly zero accuracy. These failures indicate that current LMMs are still vulnerable to adversarial OCR patterns, especially those involving visual encoding and contextual blending. In contrast, ArmorOCR achieves the best overall performance, improving over its base model by 27.3\%. The gain is particularly pronounced on AIGC Fusion Text, where ArmorOCR outperforms the strongest competing model by 65.8\%. ArmorOCR also achieves the highest average localization IoU, indicating its strong capability in both region localization and grounded adversarial OCR perception. Moreover, the mismatch between IoU and accuracy trends across different models suggests that answer accuracy alone does not guarantee reliable grounding. 
For example, GLM-5.3-Flash and Kimi-K3 achieve similar accuracy, yet GLM-5.3-Flash obtains only 40.6\% IoU, 14.3\% lower than Kimi-K3.
\textcolor{black}{ This distinction is also reflected in Joint@0.5:
GLM-5.3-Flash and Kimi-K3 achieve joint correctness rates of 15.9\% and 23.7\%, respectively, despite their similar answer accuracy. ArmorOCR achieves the highest Joint@0.5 of 46.0\%, exceeding its backbone by 21.3\% and
the strongest competing model, Gemini-3-Flash, by 17.0\%.} This highlights the importance of jointly evaluating localization and perception for fine-grained analysis of adversarial OCR failures.

\noindent\textbf{Results on Other Adversarial OCR Benchmarks.}
% The results on AdvOCR and SmuggleBench are summarized in Table~\ref{tab:other_adv}. ArmorOCR achieves the highest average accuracy on both benchmarks. 
As shown in Table~\ref{tab:other_adv}, ArmorOCR achieves the highest
average scores on both AdvOCR and SmuggleBench.
On AdvOCR, ArmorOCR outperforms the tool-assisted VACoT by 20\% on the synthetic split, which focuses on more challenging adversarial patterns.
% On AdvOCR, ArmorOCR underperforms VACoT on the real-world split, which requires fine-grained perception beyond our training objective. In contrast, ArmorOCR outperforms VACoT by 12\% on the synthetic split, which more directly targets challenging adversarial OCR patterns.
On SmuggleBench, ArmorOCR achieves better overall performance than Smuggle-CoT, with a notable gain of 8.2\% on the AI Illusions category, despite the latter relying on a much larger 235B model with detailed CoT prompting.
These results demonstrate the effectiveness and robustness of ArmorOCR across diverse adversarial OCR scenarios.
\noindent\textbf{General OCR Capability.}
A potential concern is that improving adversarial OCR perception may degrade general OCR ability. We therefore evaluate both the base model and ArmorOCR on three general OCR benchmarks. As shown in Figure~\ref{fig:general_ocr_capability}, ArmorOCR maintains comparable performance with the base model across all benchmarks, despite using no additional general OCR training data. This demonstrates that our training strategy improves adversarial OCR robustness while largely preserving general OCR capability.

\begin{table}[h]
\centering

% Left: Stage 1
\begin{minipage}[t]{0.49\linewidth}
\centering
\captionof{table}{Stage~1 component ablations.}
\label{tab:stage1_components}
\vspace{-6pt}

\scriptsize
\setlength{\tabcolsep}{2pt}
\renewcommand{\arraystretch}{1.15}
\begin{tabular*}{\linewidth}{
    @{\extracolsep{\fill}}lcccc@{\hspace{5pt}}
}
\toprule
\multicolumn{1}{c}{\raisebox{0.7ex}{\textbf{Metric}}}
& \shortstack{\textbf{Base}\\\textbf{model}}
& \shortstack{\textbf{w/o visual}\\\textbf{transfer}}
& \shortstack{\textbf{w/o region-aware}\\\textbf{distillation}}
& \shortstack{\textbf{Full}\\\textbf{Stage 1}} \\
\midrule
IoU  & 54.4 & 54.0 & 54.5 & \textbf{58.7} \\
Acc. & 29.8 & 34.7 & 44.2 & \textbf{46.8} \\
\bottomrule
\end{tabular*}
\end{minipage}
\hfill
% Right: Stage 2
\begin{minipage}[t]{0.49\linewidth}
\centering
\captionof{table}{Stage~2 reward ablations.}
\label{tab:stage2_rewards}
\vspace{-6pt}
\scriptsize
\setlength{\tabcolsep}{2pt}
\renewcommand{\arraystretch}{1.15}
\begin{tabular*}{\linewidth}{
    @{\extracolsep{\fill}}lccccc@{\hspace{5pt}}
}
\toprule
\multicolumn{1}{c}{\raisebox{0.7ex}{\textbf{Metric}}}
& \shortstack{\textbf{w/o}\\$\bm{R_{\mathrm{t2b}}}$}
& \shortstack{\textbf{w/o}\\$\bm{R_{\mathrm{b2t}}}$}
& \shortstack{\textbf{w/o}\\$\bm{R_{\mathrm{joint}}}$}
& \shortstack{\textbf{w/o}\\$\bm{R_{\mathrm{vqa}}}$}
& \shortstack{\textbf{Full}\\\textbf{Stage 2}} \\
\midrule
IoU  & 61.9 & 60.3 & 62.8 & 62.4 & \textbf{63.2} \\
Acc. & 54.8 & 51.2 & 56.6 & 53.7 & \textbf{57.1} \\
\bottomrule
\end{tabular*}
\end{minipage}

\end{table}

\vspace{-10pt}
\subsection{Ablation Study}
We conduct all ablation studies on AdvSpot, as it provides a comprehensive evaluation setting for grounded adversarial OCR perception.

% \noindent\textbf{Stage~1 Training-Method Comparison.} We compare OTSD with SFT, standard OPSD, and GRPO using the same 50K Stage~1 training examples. As shown in Table~\ref{tab:stage1_methods}, SFT, standard OPSD, and GRPO only slightly improve VQA accuracy over the base model, reaching 32.9\%, 33.9\%, and 37.3\%, respectively, compared with 46.8\% for OTSD. Together with the component ablations in Table~\ref{tab:stage1_components}, these results highlight the importance of transferring perception revealed by transformed views through response-region-aware distillation.

% \noindent\textbf{Stage~1 Training-Method Comparison.} We compare OTSD with SFT, standard OPSD, and GRPO using the same 50K Stage~1 training examples. As shown in Table~\ref{tab:stage1_methods}, SFT, standard OPSD, and GRPO improve VQA accuracy from 29.8\% for the base model to 32.9\%, 33.9\%, and 37.3\%, respectively, but all remain below OTSD at 46.8\%. Using the same training data, OTSD outperforms the alternatives, indicating that its training design contributes to the performance gains. Together with the component ablations in Table~\ref{tab:stage1_components}, these results highlight the importance of transferring perception revealed by transformed views through response-region-aware distillation.

\begin{table}[h]
\centering
\caption{\textcolor{black}{
End-to-end comparisons under identical Stage~2 GRPO refinement
on AdvSpot.
}}
\label{tab:end_to_end_comparison}
\vspace{-5pt}
\scriptsize
\setlength{\tabcolsep}{7pt}
\renewcommand{\arraystretch}{1.12}

\begin{tabular}{@{}lcccc@{}}
\toprule
\multirow{2}{*}{\textcolor{black}{\textbf{Stage 1 method}}}
& \multicolumn{2}{c}{\textcolor{black}{\textbf{Stage 1}}}
& \multicolumn{2}{c}{\textcolor{black}{\textbf{After Stage 2}}} \\
\cmidrule(lr){2-3}
\cmidrule(lr){4-5}
& \textcolor{black}{\textbf{IoU}}
& \textcolor{black}{\textbf{Acc.}}
& \textcolor{black}{\textbf{IoU}}
& \textcolor{black}{\textbf{Acc.}} \\
\midrule
SFT
& 49.2 & 32.9
& \textcolor{black}{55.8}
& \textcolor{black}{47.0} \\

OPSD + transformed views
& 54.5 & 44.2
& \textcolor{black}{59.8}
& \textcolor{black}{54.0} \\

\textbf{OTSD (ours)}
& \textbf{58.7} & \textbf{46.8}
& \textcolor{black}{\textbf{63.2}}
& \textcolor{black}{\textbf{57.1}} \\
\bottomrule
\end{tabular}
\end{table}

\noindent\textbf{Stage~1 Training-Method Comparison.} We compare OTSD with SFT, standard OPSD, and GRPO using the same 50K Stage~1 training examples. As shown in Table~\ref{tab:stage1_methods}, SFT, standard OPSD, and GRPO improve VQA accuracy from 29.8\% for the base model to 32.9\%, 33.9\%, and 37.3\%, respectively, but all remain below OTSD at 46.8\%. These results suggest that the Stage~1 gains reflect the training design, rather than exposure to synthetic data alone. The component ablations in Table~\ref{tab:stage1_components} further support the importance of transferring perception revealed by transformed views through response-region-aware distillation. \textcolor{black}{
To assess whether these advantages persist in the final system,
we further compare SFT, OPSD with transformed views, and OTSD
after identical Stage~2 GRPO refinement
(Table~\ref{tab:end_to_end_comparison}). After refinement, OTSD exceeds SFT by 7.4\%
in IoU and 10.1\% in accuracy, and exceeds OPSD with
transformed views by 3.4\% and 3.1\%, respectively.
These results support the additional contribution of
response-region-aware distillation to the final system.
}

% \noindent\textbf{Stage~1 Training Methods.} We compare OTSD with SFT, standard OPSD, and GRPO under the same Stage~1 training setting. In Table~\ref{tab:stage1_methods}, the three alternatives reach 31.5\%, 33.8\%, and 39.8\% VQA accuracy, respectively, versus 48.9\% for OTSD. Together with the component ablations in Table~\ref{tab:stage1_components}, this highlights the value of transferring perception from transformed views through response-region-aware distillation.

% \noindent\textbf{Stage Ablations.}
% We first evaluate the two-stage training design. The full ArmorOCR outperforms both single-stage variants, indicating that Stage~1 and Stage~2 are complementary. 
% Although Stage~2 alone substantially improves localization, it yields only a limited gain in VQA accuracy, highlighting the importance of \textbf{first internalizing the adversarial OCR perception revealed by transformed views}.
% Building Stage~2 upon Stage~1 further improves accuracy performance by 6.8\%, showing that task-conditioned reward optimization effectively translates the perception acquired in Stage~1 into more accurate grounded answers.

% stage 1 感觉不需要提iou
\noindent\textbf{Stage~1 Component Ablations.}
As shown in Table~\ref{tab:stage1_components}, we further analyze the two key designs in Stage~1: visual transfer and response-region-aware distillation. Removing visual transfer conditions the teacher only on the original image and privileged textual information, yielding a marginal accuracy improvement of 4.9\%, thus indicating that \textbf{privileged transformed observations provide more effective supervision than textual information alone.} Replacing our response-region-aware objective with the standard OPSD JSD loss also degrades performance, demonstrating that importance-guided selective distillation enables more effective knowledge transfer.

\noindent\textbf{Stage~2 Reward Ablations.}
As shown in Table~\ref{tab:stage2_rewards}, we finally evaluate the contribution of each task-conditioned reward in Stage~2. Removing any reward leads to a performance drop, showing that recognition, localization, spotting, and VQA objectives each provide useful supervision. Their complementary effects jointly strengthen grounded adversarial OCR perception.

\vspace{-8pt}
\section{Conclusion}
\vspace{-5pt}

In this work, we introduced \textbf{AdvSpot}, a grounded adversarial OCR benchmark with a comprehensive taxonomy for evaluating adversarial OCR perception. 
AdvSpot goes beyond existing OCR benchmarks by focusing on visually manipulated text that is readable to humans but challenging for LMMs.
We further proposed \textbf{ArmorOCR}, a two-stage framework for robust grounded adversarial OCR perception. ArmorOCR first internalizes transformation-revealed perception through observation-transferred self-distillation, and then uses task-conditioned GRPO rewards to jointly optimize localization, recognition, spotting, and grounded VQA. 
Experiments demonstrate that ArmorOCR consistently improves adversarial OCR perception while preserving general OCR capability. We hope AdvSpot and ArmorOCR will facilitate future research on reliable and generalizable visual text perception under challenging conditions.

\subsection*{AI Use Statement}
In this work, we used generative AI tools for benchmark construction and the generation and filtering of synthetic training data, as described in Section~\ref{Annotation and QA Construction} and Appendix~\ref{Training Data Construction}. We also used AI tools to polish the writing of this paper. We did not use generative AI tools for research design, experimental procedures, or analysis. We take responsibility for the final content of this work, including text and figures produced with the aid of generative AI.

% \subsection*{AI use statement}

% (This section is \textbf{required} and does not count toward the page limit.)

% In this work, we used generative AI tools for [tasks with required disclosure].
% We have not used generative AI tools for [other tasks with required disclosure],
% and [the rest of the required disclosure tasks] are not applicable to this work.
% Additionally, we used generative AI tools for [tasks with recommended
% disclosure]. We have reviewed all AI-assisted work. [Elaborate. For example, “we
% checked LLM-generated research ideas for potential plagiarism through a manual
% literature survey”, “LLM-generated code was verified and tested for correctness
% by 2 authors”, etc.]. We take responsibility for the final content of this work,
% including text, claims or artifacts produced with the aid of generative AI.

\subsection*{Ethics Statement}
We reviewed the AdvSpot images, OCR transcriptions, annotations, and question--answer pairs to ensure that the benchmark does not contain real-world personally identifiable information (PII), such as private phone numbers, physical addresses, or email addresses. We also screened the benchmark for potentially harmful or unlawful content and checked the use of its source images against applicable data licenses and privacy requirements.

% \subsection*{Ethics statement}

% (This section is \textbf{recommended} and does not count toward the page limit.)

% If authors feel that their paper submission raises questions regarding the Code
% of Ethics, they are encouraged to include a paragraph of Ethics Statement (at
% the end of the main text before references) to address potential concerns where
% appropriate. Topics include, but are not limited to, studies that involve human
% subjects, practices to data set releases, potentially harmful insights,
% methodologies and applications, potential conflicts of interest and sponsorship,
% discrimination/bias/fairness concerns, privacy and security issues, legal
% compliance, and research integrity issues (e.g., IRB, documentation, research
% ethics). 

\subsection*{Reproducibility Statement}
The main paper provides a detailed description of AdvSpot's taxonomy, annotation procedure, and evaluation protocol (Session~\ref{Benchmark: AdvSpot}), as well as the design of our ArmorOCR method (Section~\ref{Method: ArmorOCR}). The appendix provides further details on benchmark construction (Appendix~\ref{More Details of our Benchmark}), the training-data synthesis pipeline (Appendix~\ref{More Details of Training}), and the core prompts used in our experiments (Appendix~\ref{More Details of Method}). We plan to release the benchmark and code upon publication.

% \subsection*{Reproducibility statement}

% (This section is \textbf{recommended} and does not count toward the page limit.)

% It is important that the work published in ICLR is reproducible. Authors are
% strongly encouraged to include a paragraph-long Reproducibility Statement at the
% end of the main text (before references) to discuss the efforts that have been
% made to ensure reproducibility. This paragraph should not itself describe
% details needed for reproducing the results, but rather reference the parts of
% the main paper, appendix, and supplemental materials that will help with
% reproducibility. For example, for novel models or algorithms, a link to an
% anonymous downloadable source code can be submitted as supplementary materials;
% for theoretical results, clear explanations of any assumptions and a complete
% proof of the claims can be included in the appendix; for any datasets used in
% the experiments, a complete description of the data processing steps can be
% provided in the supplementary materials. Each of the above are examples of
% things that can be referenced in the reproducibility statement.



\bibliography{iclr2027_conference}
\bibliographystyle{iclr2027_conference}

\clearpage
\appendix
% \section{Appendix}
\section{More Details of our Benchmark}
\label{More Details of our Benchmark}

\begin{figure*}[h]
\centering
\centerline{\epsfig{figure=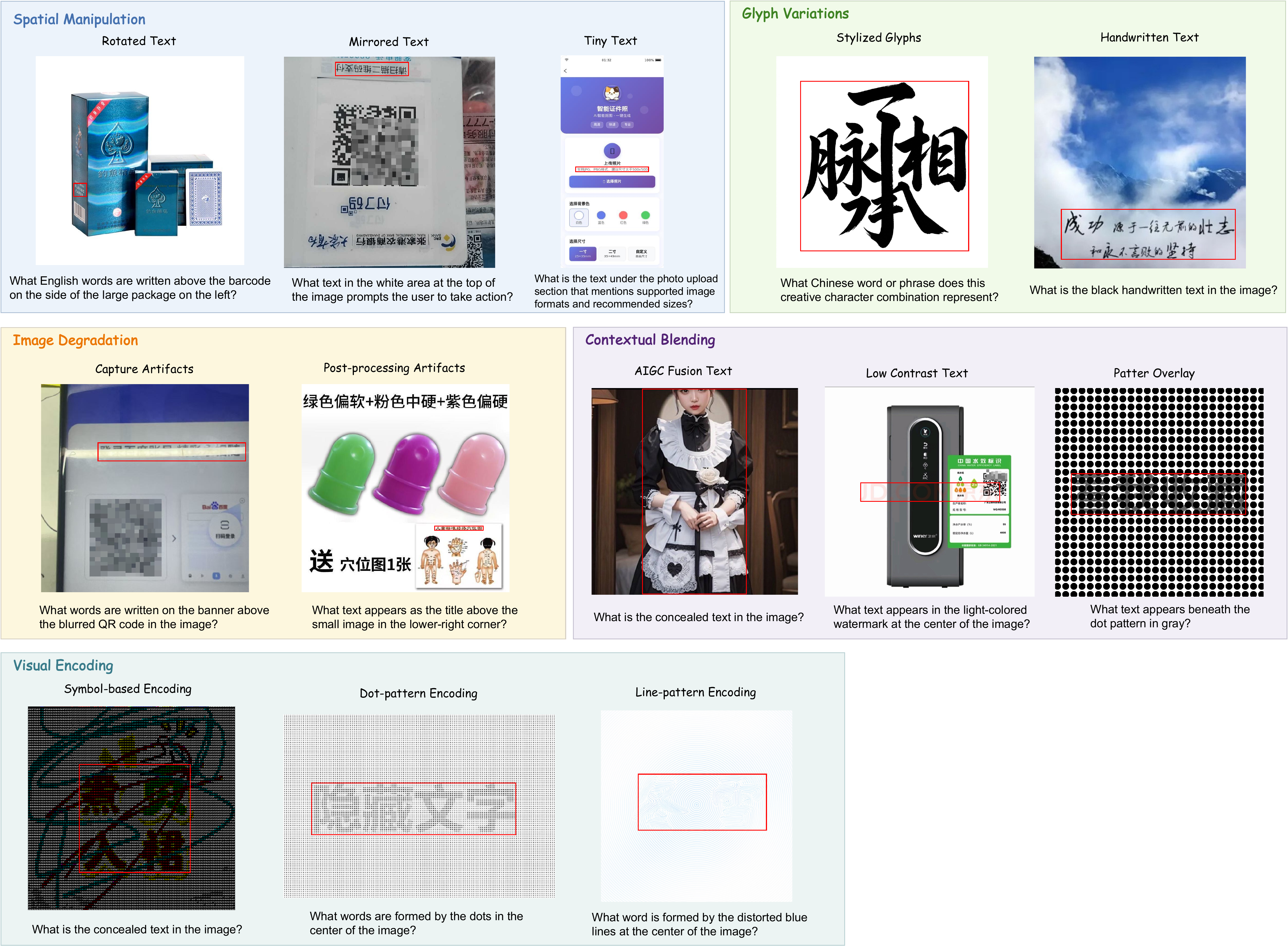,width=14cm}}
\caption{
Representative examples of the 5 primary categories and 13 fine-grained subtypes in AdvSpot. Each example includes a grounded VQA pair for region-level adversarial OCR perception evaluation.
}
\label{fig:taxonomy_examples}
\end{figure*}

\subsection{Detailed Definition of each Taxonomy}
\label{Detailed Definition of each Taxonomy}
Our taxonomy is organized according to the underlying OCR failure mechanisms. Specifically, AdvSpot categorizes adversarial OCR patterns into 5 primary categories and 13 fine-grained subtypes, as detailed below.

\begin{itemize}[leftmargin=1.5em]

\item \textbf{Spatial Manipulation.}
This category contains adversarial patterns that alter the spatial layout or geometric properties of text, making localization and recognition more challenging.

\begin{itemize}[leftmargin=1.5em]
    % \item \textbf{Rotated Text.} Text is rotated in two-dimensional or three-dimensional space, requiring the model to recognize arbitrary text orientations.
    \item \textbf{Rotated Text.}
    Text is rotated with arbitrary orientations or perspective distortions, requiring the model to recognize text under diverse geometric transformations.
    \item \textbf{Mirrored Text.} Text is horizontally or vertically flipped, violating the normal reading direction.
    \item \textbf{Tiny Text.} The text occupies only a very small portion of the image, making it difficult to localize and recognize.
\end{itemize}

\item \textbf{Glyph Variations.}
This category modifies the appearance of characters while preserving their semantic meaning.

\begin{itemize}[leftmargin=1.5em]
    \item \textbf{Stylized Glyphs.} Characters are intentionally distorted through stroke splitting, occlusion, truncation, or other artistic modifications while remaining readable to humans.
    \item \textbf{Handwritten Text.} Non-standard handwritten characters exhibiting connected strokes, exaggerated deformation, or irregular writing styles.
\end{itemize}

\item \textbf{Image Degradation.}
This category reduces OCR performance through imaging artifacts introduced during image acquisition or post-processing.

\begin{itemize}[leftmargin=1.5em]
    \item \textbf{Capture Artifacts.} Imaging degradations caused during acquisition, such as motion blur, reflections, film grain, over-/under-exposure, or excessive filtering.
    \item \textbf{Post-processing Artifacts.} Image quality degradation introduced after capture, including blur, compression artifacts, additive noise, and other image processing operations.
\end{itemize}

\item \textbf{Contextual Blending.}
This category conceals text by reducing its visual distinguishability from the surrounding context.

\begin{itemize}[leftmargin=1.5em]
    \item \textbf{AIGC Fusion Text.} AI-generated images seamlessly blend text with surrounding visual content, making text boundaries difficult to distinguish.
    \item \textbf{Low Contrast Text.} Text exhibits low contrast with its background due to similar colors or high transparency.
    \item \textbf{Pattern Overlay.} Lines, dot patterns, textures, or other visual patterns are superimposed on text, interfering with character perception.
\end{itemize}

\item \textbf{Visual Encoding.}
This category encodes textual information into non-standard visual patterns instead of conventional glyphs.

\begin{itemize}[leftmargin=1.5em]
    \item \textbf{Symbol-based Encoding.} Characters are represented through the spatial arrangement of symbols or special characters.
    \item \textbf{Dot-pattern Encoding.} Characters are encoded using the positions, sizes, or densities of dot patterns.
    \item \textbf{Line-pattern Encoding.} Characters are encoded through line orientation, length, width, or connectivity.
\end{itemize}

\end{itemize}

Figure~\ref{fig:taxonomy_examples} presents representative examples for each fine-grained subtype in our taxonomy. These examples illustrate the diverse adversarial OCR patterns covered by AdvSpot and highlight the distinct visual challenges introduced by different failure mechanisms.

Table~\ref{tab:taxonomy_statistics} summarizes the distribution of AdvSpot across different taxonomy categories and fine-grained subtypes. Counts are computed at the adversarial pattern level, where a single image or VQA instance may contain multiple adversarial patterns. AdvSpot contains 383 images with 389 grounded VQA pairs in total.

\begin{table*}[t]
\centering
\small
\renewcommand{\arraystretch}{1.1}
\setlength{\tabcolsep}{8pt}
\caption{
Distribution of AdvSpot samples across the 5 primary categories and 13 fine-grained subtypes.
}
\label{tab:taxonomy_statistics}
\begin{tabular}{llc}
\toprule
\textbf{Category} & \textbf{Subtype} & \textbf{Number of Samples} \\
\midrule

\multirow{3}{*}{Spatial Manipulation}
& Rotated Text & 35 \\
& Mirrored Text & 30 \\
& Tiny Text & 35 \\
\midrule

\multirow{2}{*}{Glyph Variations}
& Stylized Glyphs & 30 \\
& Handwritten Text & 30 \\
\midrule

\multirow{2}{*}{Image Degradation}
& Capture Artifacts & 30 \\
& Post-processing Artifacts & 30 \\
\midrule

\multirow{3}{*}{Contextual Blending}
& AIGC Fusion Text & 35 \\
& Low Contrast Text & 35 \\
& Pattern Overlay & 30 \\
\midrule

\multirow{3}{*}{Visual Encoding}
& Symbol-based Encoding & 35 \\
& Dot-pattern Encoding & 30 \\
& Line-pattern Encoding & 35 \\

\bottomrule
\end{tabular}

\end{table*}

\clearpage
\subsection{Detailed Construction of VQA Pairs}
\label{Detailed Construction of VQA Pairs}
For each retained text region, we construct a grounded visual question answering (VQA) pair using Qwen3-VL-235B-A22B-Instruct. The model receives the original image together with the target bounding box and its OCR transcription as privileged information, and is instructed to generate a question whose unique answer is exactly the provided transcription.

As shown in Figure~\ref{box:vqa_prompt}, the prompt explicitly requires the generated question to (1) uniquely identify the target region using visual attributes (e.g., object category, color, shape, size, or relative position), (2) avoid any ambiguity with other text regions in the image, (3) never expose the OCR transcription or bounding-box coordinates in the question itself, and (4) ensure that the provided transcription is the only correct answer. These constraints encourage visually grounded questions that require both accurate localization and text recognition, while preventing shortcuts based on coordinate information or answer leakage.

\begin{figure*}[h]
\centering
\begin{tcolorbox}[
    colback=gray!5, 
    colframe=gray!60!black, 
    title=\textbf{Prompt \ref{box:vqa_prompt}: System Prompt for Grounded VQA Data Construction}, 
    boxrule=0.8pt, 
    arc=2mm, 
    width=\linewidth 
]
\small\ttfamily

\textbf{You are an expert VQA data annotator.}\\
Your task is to construct Visual Question Answering (VQA) training data based on a specified text region in an image.

\textbf{Target Region:} \\
The bounding box coordinates are: [\{x\_min\}, \{y\_min\}, \{x\_max\}, \{y\_max\}].\\
The coordinate format is [x\_min, y\_min, x\_max, y\_max], normalized to the range of 0--1000 with respect to the image width and height.\\
The OCR text within this region is: \{text\_content\}.

\textbf{Your Task:}
\begin{itemize}[leftmargin=1.5em, itemsep=0.3em, parsep=0pt, label={}]
    \item \textbf{1. Visual Analysis.} \\
    Observe the target region and identify its visual characteristics, including object type, color, shape, size, and relative position.

    \item \textbf{2. Question Construction.} \\
    Construct a question asking about the text content within the target region.

    \item \textbf{3. Unique Grounding Requirement.} \\
    The question must uniquely identify the target region in the entire image and should not introduce ambiguity with other regions.

    \item \textbf{4. Content Constraints.} \\
    The question must not contain coordinate values, coordinate numbers, or the OCR text itself. The only correct answer to the question must exactly match the provided OCR text.
\end{itemize}

\vspace{0.3em}
\begin{tcolorbox}[colback=white, colframe=gray!30, title=\textbf{Output Format Requirement}, boxrule=0.5pt, sharp corners, left=2pt, right=2pt, top=2pt, bottom=2pt]
\footnotesize
Question: [Your generated question]
\end{tcolorbox}

\textbf{Important Notes:}\\
The question must never reveal the provided OCR text or use coordinates to refer to the target region. The OCR text is only provided to understand the semantic role of the target region and its relationship with other visual elements in the image.

\end{tcolorbox}

\caption{The system prompt used for constructing grounded VQA training data.}
\label{box:vqa_prompt}
\end{figure*}

% \noindent\textbf{Prompt:}

% \noindent\textbf{Why construct VQA:}

% \noindent\textbf{Characteristics of AdvSpot}

% \subsection{Detailed Testing Prompt}

% Figure~\ref{box:test_prompt} presents the unified testing prompt used throughout all experiments. Given a grounded VQA question, the model is instructed to first localize the referenced text region and then recognize its text content, returning both the predicted bounding box and answer in a predefined format.

% \begin{figure*}[t]
% \centering
% \begin{tcolorbox}[
%     colback=gray!5, 
%     colframe=gray!60!black, 
%     title=\textbf{Prompt 2: Detailed Testing Prompt for Grounded OCR Perception Evaluation}, 
%     boxrule=0.8pt, 
%     arc=2mm, 
%     width=\linewidth 
% ]
% \small\ttfamily

% \textbf{Question:}\\
% \{question\}\\

% Please output the answer and the corresponding text bounding box in JSON format:
% \texttt{\{"text":"answer content","bbox":[x\_min,y\_min,x\_max,y\_max]\}}.\\
% \texttt{The coordinate values are in the range of 0--1000.}

% \end{tcolorbox}

% \caption{The detailed testing prompt used for grounded OCR perception evaluation.}
% \label{box:test_prompt}
% \end{figure*}

\subsection{Detailed Testing Prompt}
\label{Detailed Testing Prompt}

% Figure~\ref{box:test_prompt} presents the bilingual (Chinese--English) testing prompt used throughout all experiments. Given a grounded VQA question, the model is instructed to output the recognized text and its corresponding bounding box in a predefined JSON format. The bounding box coordinates are normalized to the range of 0--1000 with respect to the image size.

% To ensure stable and reproducible benchmarking, we adopt the Chinese instruction as the unified evaluation prompt for all experiments. The English version is provided for clarity and facilitates understanding of the evaluation protocol.

Figure~\ref{box:test_prompt} presents the testing prompt used throughout all experiments. Given a grounded VQA question, the model is instructed to output the recognized text and its corresponding bounding box in a predefined JSON format. The bounding box coordinates are normalized to the range of 0--1000 with respect to the image size.

% To ensure stable and reproducible benchmarking, we adopt the Chinese instruction as the unified evaluation prompt for all experiments. The English version is provided for clarity and facilitates understanding of the evaluation protocol.

\begin{figure*}[t]
\centering
\begin{tcolorbox}[
    colback=gray!5, 
    colframe=gray!60!black, 
    title=\textbf{Prompt 2: Detailed Testing Prompt for Grounded OCR Perception Evaluation}, 
    boxrule=0.8pt, 
    arc=2mm, 
    width=\linewidth 
]
\small\ttfamily

Please output the answer and the corresponding text bounding box in JSON format:\\
\texttt{\{"text":"answer content","bbox":[x\_min,y\_min,x\_max,y\_max]\}}\\
The coordinate values are in the range of 0--1000.

\end{tcolorbox}

\caption{
The testing prompt used for grounded OCR perception evaluation. 
}
\label{box:test_prompt}
\end{figure*}

\begin{table*}[t!]
\centering
\caption{\textbf{Localization IOU} results of our AdvSpot benchmark across the expanded model zoo. 
\textbf{Category legend:}
\textbf{1} = Imaging Degradation (1-1: Capture Artifacts; 1-2: Post-processing Artifacts);
\textbf{2} = Spatial Manipulation (2-1: Rotated Text; 2-2: Mirrored Text; 2-3: Tiny Text);
\textbf{3} = Glyph Variations (3-1: Stylized Glyphs; 3-2: Handwritten Text);
\textbf{4} = Visual Encoding (4-1: Symbol-based Encoding; 4-2: Dot-pattern Encoding; 4-3: Line-pattern Encoding);
\textbf{5} = Contextual Blending (5-1: AIGC Fusion Text; 5-2: Low Contrast Text; 5-3: Pattern Overlay).
}
\label{tab:final_models_iou}

\begingroup
\fontsize{8pt}{9.2pt}\selectfont
\setlength{\tabcolsep}{1.8pt}
\renewcommand{\arraystretch}{1.08}
\setlength{\belowrulesep}{1pt}
\setlength{\aboverulesep}{1pt}

\begin{tabularx}{\textwidth}{@{}l*{14}{>{\centering\arraybackslash}X}@{}}
\toprule
\textbf{Model}
& \multicolumn{2}{c}{\textbf{1}}
& \multicolumn{3}{c}{\textbf{2}}
& \multicolumn{2}{c}{\textbf{3}}
& \multicolumn{3}{c}{\textbf{4}}
& \multicolumn{3}{c}{\textbf{5}}
& \multicolumn{1}{c}{\textbf{Overall}} \\
\cmidrule(lr){2-3}
\cmidrule(lr){4-6}
\cmidrule(lr){7-8}
\cmidrule(lr){9-11}
\cmidrule(lr){12-14}
& 1-1 & 1-2
& 2-1 & 2-2 & 2-3
& 3-1 & 3-2
& 4-1 & 4-2 & 4-3
& 5-1 & 5-2 & 5-3
& \\
\midrule

\rowcolor{gray!10}
\multicolumn{15}{@{}l}{\textit{\textbf{OpenAI GPT Family}}} \\
% GPT-4o
% & 5.3 & 11.6 & 3.5 & 8.1 & 4.3
% & 26.0 & 17.1 & 27.8 & 30.0 & 19.4
% & 8.8 & 13.3 & 23.3 & 15.9 \\
% GPT-5
% & 16.0 & 22.8 & 19.9 & 19.6 & 11.7
% & 55.8 & 40.0 & 54.8 & 51.9 & 48.2
% & 20.4 & 23.6 & 52.1 & 35.0 \\
GPT-5.1           & 10.5 & 18.8 & 11.5 & 13.8 & 3.8 & 40.6 & 37.2 & 44.7 & 50.0 & 42.9 & 28.1 & 7.5 & 38.7 & 27.8 \\
% GPT-5.4
% & 20.8 & 27.8 & 22.5 & 27.4 & 11.0
% & 56.3 & 37.0 & 61.1 & 46.9 & 54.1
% & 25.4 & 14.0 & 38.4 & 34.5 \\
GPT-5.5           & 39.7 & 42.5 & 34.7 & 42.4 & 18.7 & 62.9 & 59.6 & 78.7 & 78.8 & 68.4 & 33.1 & 37.5 & 69.2 & 51.5 \\
GPT-5.6-sol       & 59.3 & 56.7 & 52.6 & 66.1 & 46.0 & 72.9 & 76.4 & 81.0 & 80.9 & 72.2 & 47.8 & 55.7 & 69.8 & 64.2 \\
\midrule

\rowcolor{gray!10}
\multicolumn{15}{@{}l}{\textit{\textbf{Google Gemini Family}}} \\
% Gemini-2.5-Flash  & -- & -- & -- & -- & -- & -- & -- & -- & -- & -- & -- & -- & -- & -- \\
% Gemini-2.5-Pro    & -- & -- & -- & -- & -- & -- & -- & -- & -- & -- & -- & -- & -- & -- \\
Gemini-3-Flash    & 45.0 & 56.5 & 43.5 & 45.7 & 39.6 & 66.7 & 54.0 & 69.5 & 77.5 & 80.7 & 52.4 & 37.5 & 91.0 & 59.2 \\
% Gemini-3.5-Flash  & -- & -- & -- & -- & -- & -- & -- & -- & -- & -- & -- & -- & -- & -- \\
Gemini-3.6-Flash  & 31.4 & 33.4 & 22.8 & 40.3 & 22.6 & 62.2 & 54.9 & 68.0 & 82.8 & 80.0 & 53.7 & 19.0 & 90.5 & 52.0 \\
\midrule

\rowcolor{gray!10}
\multicolumn{15}{@{}l}{\textit{\textbf{Anthropic Claude Family}}} \\
Claude-Haiku-4.5
& 3.7 & 7.2 & 1.1 & 6.0 & 6.0
& 22.2 & 6.3 & 16.5 & 21.2 & 24.5
& 16.1 & 6.3 & 22.0 & 12.8 \\
Claude-Sonnet-4.5
& 4.8 & 9.7 & 10.3 & 12.1 & 0.0
& 32.3 & 19.8 & 3.2 & 4.0 & 11.7
& 5.8 & 6.0 & 12.9 & 10.4 \\
Claude-Opus-4.5
& 4.0 & 8.1 & 4.1 & 7.8 & 4.7
& 12.5 & 13.7 & 16.3 & 19.6 & 26.0
& 16.7 & 5.1 & 25.1 & 13.1 \\
\midrule

\rowcolor{gray!10}
\multicolumn{15}{@{}l}{\textit{\textbf{Moonshot Family}}} \\
Kimi-K2.5         & 50.5 & 41.7 & 44.9 & 44.6 & 42.0 & 74.6 & 60.2 & 43.1 & 54.9 & 46.9 & 29.1 & 33.4 & 56.4 & 47.3 \\
Kimi-K2.6         & 49.1 & 40.5 & 52.0 & 46.7 & 40.3 & 75.8 & 65.5 & 45.5 & 60.1 & 51.8 & 30.9 & 34.1 & 57.5 & 49.3 \\
Kimi-K3           & 50.9 & 49.6 & 61.6 & 55.0 & 47.3 & 83.3 & 73.9 & 50.7 & 64.6 & 49.6 & 39.5 & 40.6 & 51.3 & 54.9 \\
\midrule

\rowcolor{gray!10}
\multicolumn{15}{@{}l}{\textit{\textbf{Zhipu Family}}} \\
% GLM-4.5V          & -- & -- & -- & -- & -- & -- & -- & -- & -- & -- & -- & -- & -- & -- \\
GLM-4.6V          & 47.1 & 44.0 & 49.9 & 53.6 & 31.6 & 73.7 & 74.7 & 45.4 & 78.4 & 70.5 & 36.9 & 46.1 & 75.1 & 56.1 \\
GLM-5.3-Flash     & 30.3 & 33.2 & 28.8 & 33.1 & 13.0 & 62.9 & 51.8 & 44.2 & 56.7 & 61.5 & 22.6 & 26.9 & 56.0 & 40.6 \\
\midrule

\rowcolor{gray!10}
\multicolumn{15}{@{}l}{\textit{\textbf{Alibaba Qwen Family}}} \\
\rowcolor{gray!5}
\multicolumn{15}{@{}l}{\textit{\underline{Qwen3 Series}}} \\
Qwen3-VL-8B
& 59.1 & 43.3 & 68.0 & 65.7 & 45.9
& 76.2 & 71.2 & 28.7 & 65.5 & 43.5
& 38.1 & 52.4 & 62.3 & 54.4 \\
Qwen3-VL-30B-A3B
& 61.0 & 45.9 & 69.0 & 73.6 & 46.8
& 73.1 & 71.4 & 39.8 & 60.4 & 56.4
& 21.3 & 47.6 & 62.0 & 55.0 \\
Qwen3-VL-32B
& 70.5 & 55.0 & 69.7 & 72.0 & 60.8
& 79.6 & 76.8 & 50.8 & 63.1 & 53.2
& 42.5 & 49.8 & 66.8 & 61.1 \\
Qwen3-VL-235B-A22B & 59.7 & 46.8 & 71.7 & 68.3 & 48.9 & 74.6 & 75.7 & 50.2 & 61.4 & 53.7 & 38.3 & 54.2 & 67.6 & 58.6 \\
\rowcolor{gray!5}
\multicolumn{15}{@{}l}{\textit{\underline{Qwen3.5 Series}}} \\
Qwen3.5-35B-A3B & 39.2 & 35.7 & 54.2 & 44.2 & 26.5 & 67.8 & 56.4 & 41.8 & 52.0 & 45.0 & 31.4 & 35.2 & 44.8 & 43.9 \\
Qwen3.5-122B-A10B & 50.2 & 43.7 & 51.6 & 51.0 & 44.6 & 61.2 & 57.9 & 25.7 & 48.1 & 44.0 & 19.7 & 49.2 & 58.3 & 45.8 \\
Qwen3.5-397B-A17B & 29.2 & 28.6 & 33.8 & 32.6 & 12.0 & 58.8 & 44.5 & 41.8 & 53.8 & 49.4 & 28.7 & 27.2 & 42.2 & 37.3 \\
\midrule
\rowcolor{gray!5}
\multicolumn{15}{@{}l}{\textit{\underline{Qwen3.8 Series}}} \\
Qwen3.8-27B & 57.1 & 40.6 & 59.2 & 66.3 & 40.6 & 73.7 & 56.9 & 43.7 & 64.6 & 50.8 & 22.2 & 42.9 & 60.4 & 51.2 \\
\midrule

\rowcolor{gray!10}
\multicolumn{15}{@{}l}{\textit{\textbf{Our Proposed Method}}} \\
ArmorOCR
& 58.9 & 40.3 & 70.3 & 60.9 & 41.2
& 70.1 & 61.0 & 72.4 & 73.0 & 77.0
& 61.5 & 41.0 & 89.1 & 63.2 \\
\bottomrule
\end{tabularx}
\endgroup
\end{table*}

\begin{table*}[t!]
\centering
\caption{\textbf{VQA Accuracy} results of our AdvSpot benchmark across the expanded model zoo.
}
\label{tab:final_models_acc}

\begingroup
\fontsize{8pt}{9.2pt}\selectfont
\setlength{\tabcolsep}{1.8pt}
\renewcommand{\arraystretch}{1.08}
\setlength{\belowrulesep}{1pt}
\setlength{\aboverulesep}{1pt}

\begin{tabularx}{\textwidth}{@{}l*{14}{>{\centering\arraybackslash}X}@{}}
\toprule
\textbf{Model}
& \multicolumn{2}{c}{\textbf{1}}
& \multicolumn{3}{c}{\textbf{2}}
& \multicolumn{2}{c}{\textbf{3}}
& \multicolumn{3}{c}{\textbf{4}}
& \multicolumn{3}{c}{\textbf{5}}
& \multicolumn{1}{c}{\textbf{Overall}} \\
\cmidrule(lr){2-3}
\cmidrule(lr){4-6}
\cmidrule(lr){7-8}
\cmidrule(lr){9-11}
\cmidrule(lr){12-14}
& 1-1 & 1-2
& 2-1 & 2-2 & 2-3
& 3-1 & 3-2
& 4-1 & 4-2 & 4-3
& 5-1 & 5-2 & 5-3
& \\
\midrule

\rowcolor{gray!10}
\multicolumn{15}{@{}l}{\textit{\textbf{OpenAI GPT Family}}} \\
% GPT-4o
% & 20.0 & 43.3 & 26.7 & 13.3 & 50.0
% & 16.7 & 33.3 & 20.0 & 20.0 & 20.0
% & 0.5 & 31.4 & 5.7 & 23.2 \\
% GPT-5
% & 30.0 & 40.0 & 30.0 & 23.3 & 60.0
% & 43.3 & 40.0 & 32.5 & 26.7 & 20.0
% & 2.5 & 34.4 & 8.6 & 30.0 \\
GPT-5.1           & 26.7 & 43.3 & 28.6 & 13.3 & 45.7 & 30.0 & 30.0 & 31.4 & 33.3 & 11.4 & 5.7 & 28.6 & 13.3 & 26.5 \\
% GPT-5.4
% & 23.3 & 46.7 & 23.3 & 13.3 & 36.7
% & 13.3 & 20.0 & 0.0 & 3.3 & 3.3
% & 0.0 & 28.6 & 5.7 & 16.1 \\
GPT-5.5           & 26.7 & 40.0 & 28.6 & 20.0 & 51.4 & 50.0 & 46.7 & 2.9 & 10.0 & 2.9 & 2.9 & 37.1 & 6.7 & 24.9 \\
GPT-5.6-sol       & 23.3 & 33.3 & 40.0 & 20.0 & 51.4 & 36.7 & 40.0 & 2.9 & 10.0 & 2.9 & 0.0 & 37.1 & 3.3 & 22.9 \\
\midrule

\rowcolor{gray!10}
\multicolumn{15}{@{}l}{\textit{\textbf{Google Gemini Family}}} \\
% Gemini-2.5-Flash   & -- & -- & -- & -- & -- & -- & -- & -- & -- & -- & -- & -- & -- & -- \\
% Gemini-2.5-Pro     & -- & -- & -- & -- & -- & -- & -- & -- & -- & -- & -- & -- & -- & -- \\
Gemini-3-Flash    & 66.7 & 80.0 & 62.9 & 13.3 & 68.6 & 56.7 & 56.7 & 20.0 & 33.3 & 11.4 & 17.1 & 74.3 & 36.7 & 45.2 \\
% Gemini-3.5-Flash   & -- & -- & -- & -- & -- & -- & -- & -- & -- & -- & -- & -- & -- & -- \\
Gemini-3.6-Flash  & 73.3 & 80.0 & 65.7 & 23.3 & 77.1 & 63.3 & 66.7 & 8.6 & 40.0 & 17.1 & 8.6 & 74.3 & 33.3 & 47.6 \\
\midrule

\rowcolor{gray!10}
\multicolumn{15}{@{}l}{\textit{\textbf{Anthropic Claude Family}}} \\
Claude-Haiku-4.5
& 3.3 & 16.7 & 3.3 & 0.0 & 26.7
& 6.7 & 3.3 & 0.0 & 0.0 & 0.0
& 0.0 & 8.6 & 0.0 & 5.0 \\
Claude-Sonnet-4.5
& 3.3 & 26.7 & 6.7 & 6.7 & 50.0
& 13.3 & 13.3 & 0.0 & 3.3 & 0.0
& 0.0 & 11.4 & 0.0 & 9.8 \\
Claude-Opus-4.5
& 33.3 & 50.0 & 30.0 & 13.3 & 60.0
& 30.0 & 26.7 & 0.0 & 3.3 & 3.3
& 5.0 & 31.4 & 2.9 & 21.2 \\
\midrule

\rowcolor{gray!10}
\multicolumn{15}{@{}l}{\textit{\textbf{Moonshot Family}}} \\
Kimi-K2.5          & 50.0 & 63.3 & 45.7 & 13.3 & 60.0 & 76.7 & 56.7 & 8.6 & 13.3 & 2.9 & 8.6 & 54.3 & 10.0 & 34.4 \\
Kimi-K2.6          & 53.3 & 60.0 & 45.7 & 20.0 & 62.9 & 73.3 & 70.0 & 11.4 & 13.3 & 0.0 & 5.7 & 54.3 & 16.7 & 37.0 \\
Kimi-K3            & 53.3 & 56.7 & 51.4 & 30.0 & 68.6 & 70.0 & 60.0 & 2.9 & 13.3 & 2.9 & 5.7 & 48.6 & 3.3 & 35.0 \\
\midrule

\rowcolor{gray!10}
\multicolumn{15}{@{}l}{\textit{\textbf{Zhipu Family}}} \\
%GLM-4.5V           & -- & -- & -- & -- & -- & -- & -- & -- & -- & -- & -- & -- & -- & -- \\
GLM-4.6V           & 50.0 & 36.7 & 31.4 & 23.3 & 51.4 & 36.7 & 36.7 & 5.7 & 23.3 & 5.7 & 14.3 & 28.6 & 33.3 & 30.3 \\
GLM-5.3-Flash      & 50.0 & 40.0 & 48.6 & 43.3 & 45.7 & 60.0 & 53.3 & 5.7 & 26.7 & 0.0 & 11.4 & 45.7 & 36.7 & 35.2 \\
\midrule

\rowcolor{gray!10}
\multicolumn{15}{@{}l}{\textit{\textbf{Alibaba Qwen Family}}} \\
\rowcolor{gray!5}
\multicolumn{15}{@{}l}{\textit{\underline{Qwen3 Series}}} \\
Qwen3-VL-8B
& 40.0 & 53.3 & 48.6 & 20.0 & 60.0
& 30.0 & 36.7 & 0.0 & 23.3 & 2.9
& 5.7 & 45.7 & 26.7 & 29.8 \\
Qwen3-VL-30B-A3B
& 53.3 & 53.3 & 48.6 & 26.7 & 54.3
& 50.0 & 53.3 & 0.0 & 23.3 & 5.7
& 8.6 & 45.7 & 30.0 & 34.7 \\
Qwen3-VL-32B
& 53.3 & 50.0 & 54.3 & 20.0 & 51.4
& 46.7 & 40.0 & 0.0 & 20.0 & 5.7
& 5.7 & 48.6 & 23.3 & 31.9 \\
Qwen3-VL-235B-A22B & 63.3 & 60.0 & 60.0 & 23.3 & 57.1 & 53.3 & 53.3 & 5.7 & 23.3 & 14.3 & 11.4 & 48.6 & 26.7 & 38.3 \\
\rowcolor{gray!5}
\multicolumn{15}{@{}l}{\textit{\underline{Qwen3.5 Series}}} \\
Qwen3.5-35B-A3B & 43.3 & 53.3 & 48.6 & 20.0 & 40.0 & 43.3 & 50.0 & 8.6 & 20.0 & 0.0 & 8.6 & 34.3 & 26.7 & 30.1 \\
Qwen3.5-122B-A10B & 23.3 & 33.3 & 34.3 & 10.0 & 48.6 & 46.7 & 26.7 & 0.0 & 10.0 & 0.0 & 2.9 & 34.3 & 23.3 & 22.1 \\
Qwen3.5-397B-A17B & 33.3 & 36.7 & 28.6 & 6.7 & 31.4 & 46.7 & 20.0 & 2.9 & 16.7 & 0.0 & 0.0 & 25.7 & 13.3 & 19.5 \\
\midrule
\rowcolor{gray!5}
\multicolumn{15}{@{}l}{\textit{\underline{Qwen3.8 Series}}} \\
Qwen3.8-27B & 46.7 & 43.3 & 42.9 & 13.3 & 48.6 & 36.7 & 26.7 & 0.0 & 16.7 & 8.6 & 0.0 & 37.1 & 16.7 & 25.4 \\
\midrule

\rowcolor{gray!10}
\multicolumn{15}{@{}l}{\textit{\textbf{Our Proposed Method}}} \\
ArmorOCR
& 43.3 & 60.0 & 54.3 & 46.7 & 57.1
& 26.7 & 46.7 & 68.6 & 63.3 & 54.3
& 82.9 & 42.9 & 66.7 & 57.1 \\
\bottomrule
\end{tabularx}
\endgroup
\end{table*}
% \clearpage

\subsection{More Experiments on AdvSpot}
\label{More Experiments on AdvSpot}

\medskip
\noindent\textcolor{black}{\textbf{Extended Model Evaluation.}}
To provide a more comprehensive evaluation, we report the complete benchmarking results on AdvSpot across an expanded collection of both open-source and proprietary LMMs. Specifically, Table~\ref{tab:final_models_iou} presents the localization IoU results, while Table~\ref{tab:final_models_acc} reports the corresponding grounded VQA accuracy.

\medskip
\noindent\textcolor{black}{\textbf{Cross-benchmark Comparisons.}}
\textcolor{black}{We compare the same nine model versions across AdvOCR, SmuggleBench, and AdvSpot. Table~\ref{tab:cross_benchmark_comparison} reports each model's score and rank within this comparison set. Model rankings vary across benchmarks. For example, GLM-5.3-Flash outperforms Gemini-3-Flash on AdvOCR (38.0\% vs.\ 30.0\%), but trails it on SmuggleBench TER (16.4\% vs.\ 18.4\%) and AdvSpot accuracy (35.2\% vs.\ 45.2\%). These rank reversals suggest that the benchmarks capture different aspects of model performance. Within AdvSpot, answer accuracy and joint correctness also produce different rankings. GLM-5.3-Flash ranks fourth in accuracy (35.2\%) but eighth in Joint@0.5 (15.9\%), whereas Qwen3-VL-30B-A3B ranks sixth in accuracy (34.7\%) but ties for third in Joint@0.5 (28.8\%). These differences illustrate the complementary information provided by evaluating joint answer and localization correctness.}

\begin{table}[ht]
\centering
\caption{\textcolor{black}{Cross-benchmark comparison of the same nine model versions. Each entry reports score (\%) / rank among the nine models.}}
\label{tab:cross_benchmark_comparison}
\vspace{-3pt}
\scriptsize
\setlength{\tabcolsep}{5pt}
\renewcommand{\arraystretch}{1.12}

\begin{tabular}{@{}lcccc@{}}
\toprule
\textcolor{black}{\textbf{Model}}
& \textcolor{black}{\shortstack{\textbf{AdvOCR}\\\textbf{Avg. Acc.}}}
& \textcolor{black}{\shortstack{\textbf{SmuggleBench}\\\textbf{TER}}}
& \textcolor{black}{\shortstack{\textbf{AdvSpot}\\\textbf{Acc.}}}
& \textcolor{black}{\shortstack{\textbf{AdvSpot}\\\textbf{Joint@0.5}}} \\
\midrule
Qwen3-VL-8B
& 21.0 \textcolor{black}{/ 8}
& 13.3 \textcolor{black}{/ 8}
& 29.8 \textcolor{black}{/ 8}
& \textcolor{black}{24.7 / 7} \\

Qwen3-VL-30B-A3B
& 23.0 \textcolor{black}{/ 7}
& 14.8 \textcolor{black}{/ 5}
& 34.7 \textcolor{black}{/ 6}
& \textcolor{black}{28.8 / 3} \\

Qwen3-VL-32B
& 26.0 \textcolor{black}{/ 6}
& 14.3 \textcolor{black}{/ 7}
& 31.9 \textcolor{black}{/ 7}
& \textcolor{black}{28.8 / 3} \\

Qwen3-VL-235B-A22B
& 28.0 \textcolor{black}{/ 5}
& 14.6 \textcolor{black}{/ 6}
& 38.3 \textcolor{black}{/ 3}
& \textcolor{black}{28.0 / 5} \\

GPT-5.1
& \textcolor{black}{20.0 / 9}
& \textcolor{black}{7.2 / 9}
& 26.5 \textcolor{black}{/ 9}
& \textcolor{black}{6.7 / 9} \\

Gemini-3-Flash
& \textcolor{black}{30.0 / 4}
& \textcolor{black}{\textbf{18.4 / 1}}
& 45.2 \textcolor{black}{/ 2}
& \textcolor{black}{29.0 / 2} \\

GLM-5.3-Flash
& \textcolor{black}{38.0 / 2}
& \textcolor{black}{16.4 / 3}
& 35.2 \textcolor{black}{/ 4}
& \textcolor{black}{15.9 / 8} \\

Kimi-K3
& \textcolor{black}{31.0 / 3}
& \textcolor{black}{16.1 / 4}
& 35.0 \textcolor{black}{/ 5}
& \textcolor{black}{23.7 / 6} \\

\textbf{ArmorOCR}
& \textbf{56.0} \textcolor{black}{\textbf{/ 1}}
& 17.1 \textcolor{black}{/ 2}
& \textbf{57.1} \textcolor{black}{\textbf{/ 1}}
& \textcolor{black}{\textbf{46.0 / 1}} \\
\bottomrule
\end{tabular}
\end{table}

% 导言区需要：
% \usepackage{tabularx}
% \usepackage{array}
% \usepackage{booktabs}
% \usepackage[table]{xcolor}

\begin{figure}[ht]
\centering
\includegraphics[width=\linewidth]{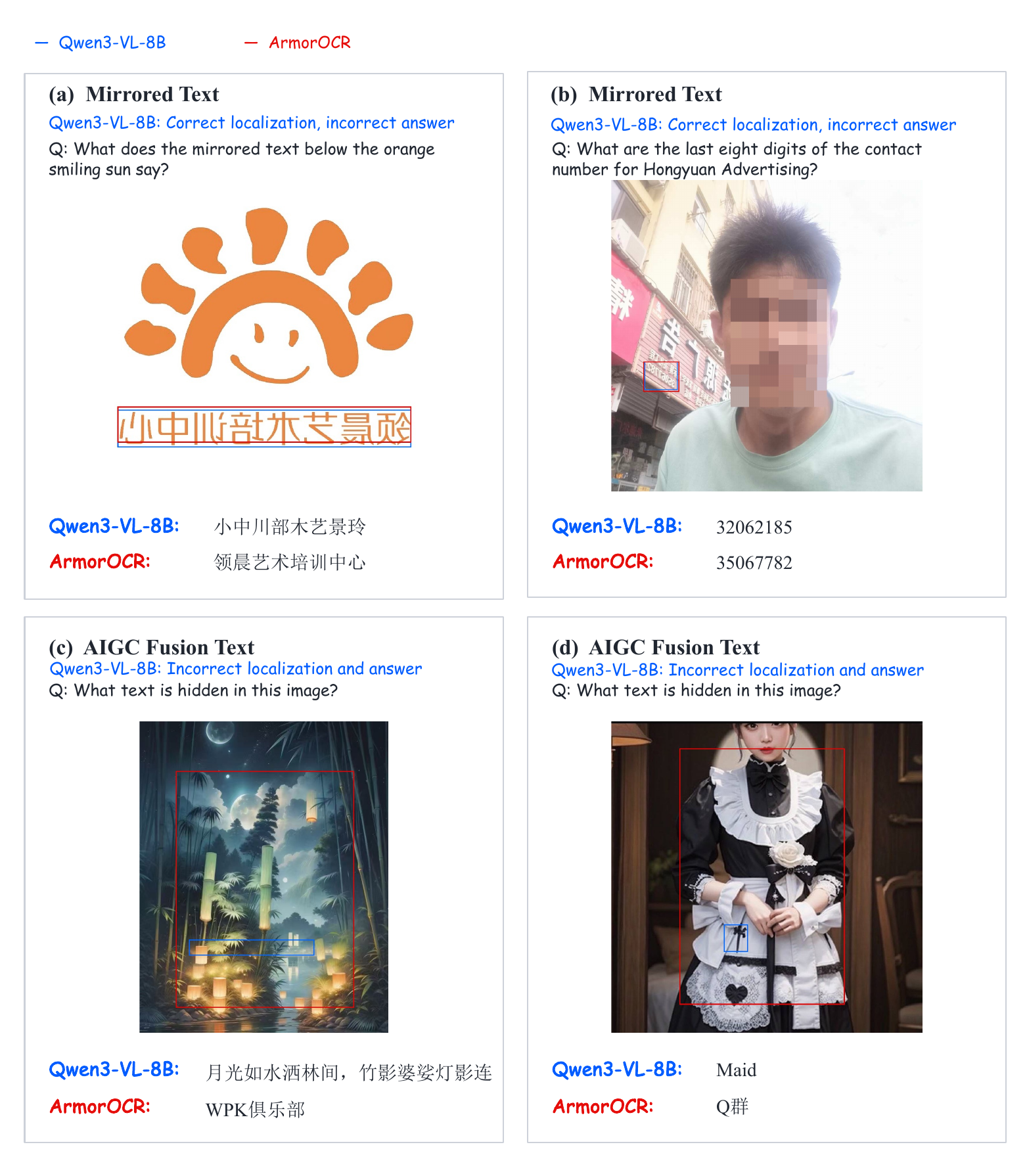}
\caption{\textcolor{black}{Qualitative examples of distinct failure modes on AdvSpot. We show the question and predictions from Qwen3-VL-8B and ArmorOCR. These examples distinguish incorrect answers with correct localization from failures in both answer and localization.}}
\label{fig:advspot_failure_cases}
\end{figure}

\medskip
\noindent\textcolor{black}{\textbf{Qualitative Failure Analysis.}}
\textcolor{black}{Figure~\ref{fig:advspot_failure_cases} illustrates how AdvSpot distinguishes failures that answer-only scoring groups together. An incorrect answer may occur despite correct localization, or alongside localization of an incorrect region. By examining predicted answers and bounding boxes together, AdvSpot separates these outcomes and provides a more detailed diagnosis of adversarial OCR failures.}

\medskip
\noindent\textcolor{black}{\textbf{Statistical Reliability.}}
\textcolor{black}{We assess the overall performance differences between ArmorOCR and Qwen3-VL-8B on AdvSpot using 10,000 image-level paired bootstrap resamples. In each resample, we sample images with replacement, use the same sampled images for both models, and keep all QA instances associated with each image together. We recompute the aggregate metrics and their paired differences for each resample. The 95\% confidence intervals are given by the 2.5th and 97.5th percentiles of the bootstrap differences. As shown in Table~\ref{tab:advspot_bootstrap}, the 95\% confidence intervals remain entirely above zero across all three metrics, supporting AdvSpot's reliability in distinguishing the overall model performance.}

\begin{table}[h]
\centering
\caption{\textcolor{black}{Overall performance differences and bootstrap confidence intervals on AdvSpot. $\Delta$ denotes ArmorOCR minus Qwen3-VL-8B.}}
\label{tab:advspot_bootstrap}
\vspace{-3pt}
\scriptsize
\setlength{\tabcolsep}{7pt}
\renewcommand{\arraystretch}{1.12}

\begin{tabular}{@{}lccc@{}}
\toprule
\textcolor{black}{\textbf{Model / Statistic}}
& \textcolor{black}{\textbf{Acc.}}
& \textcolor{black}{\textbf{IoU}}
& \textcolor{black}{\textbf{Joint@0.5}} \\
\midrule
Qwen3-VL-8B
& 29.8
& 54.4
& \textcolor{black}{24.7} \\

ArmorOCR
& 57.1
& 63.2
& \textcolor{black}{46.0} \\
\midrule
\textcolor{black}{\textbf{$\Delta$ (ArmorOCR $-$ Qwen3-VL-8B)}}
& \textcolor{black}{\textbf{+27.3}}
& \textcolor{black}{\textbf{+8.8}}
& \textcolor{black}{\textbf{+21.3}} \\

\textcolor{black}{\textbf{95\% bootstrap CI for $\Delta$}}
& \textcolor{black}{\textbf{[22.1, 32.8]}}
& \textcolor{black}{\textbf{[4.1, 11.7]}}
& \textcolor{black}{\textbf{[14.8, 25.3]}} \\
\bottomrule
\end{tabular}
\end{table}

\medskip
\noindent\textcolor{black}{\textbf{Localization Ability on Clean versus Adversarial Images.}}
\textcolor{black}{We evaluate Qwen3-VL-8B on 500 matched clean--adversarial image pairs across five mechanisms, with 100 pairs per mechanism. Each pair shares the same text content, layout, and ground-truth bounding box, differing only in the adversarial presentation. We report both answer accuracy and mean localization IoU to examine how adversarial presentation affects recognition and localization.}

\begin{table}[ht]
\centering
\caption{\textcolor{black}{Performance of Qwen3-VL-8B on matched clean and adversarial images. Each mechanism contains 100 image pairs. Accuracy and mean IoU are reported in \%. Drop denotes the clean score minus the adversarial score.}}
\label{tab:clean_adversarial_comparison}
\vspace{-3pt}
\scriptsize
\setlength{\tabcolsep}{6pt}
\renewcommand{\arraystretch}{1.12}

{\color{black}
\begin{tabular}{@{}lcccccc@{}}
\toprule
\multirow{2}{*}{\textbf{Mechanism}}
& \multicolumn{3}{c}{\textbf{Accuracy}}
& \multicolumn{3}{c}{\textbf{Mean IoU}} \\
\cmidrule(lr){2-4}
\cmidrule(lr){5-7}
& \textbf{Clean}
& \textbf{Adv.}
& \textbf{Drop}
& \textbf{Clean}
& \textbf{Adv.}
& \textbf{Drop} \\
\midrule
Mirrored Text
& 99.0 & 2.0 & 97.0
& 92.51 & 92.12 & 0.39 \\

Pattern Overlay
& 100.0 & 74.0 & 26.0
& 91.75 & 90.51 & 1.24 \\

Dot-pattern Encoding
& 100.0 & 2.0 & 98.0
& 91.98 & 72.14 & 19.84 \\

Line-pattern Encoding
& 99.0 & 27.0 & 72.0
& 92.18 & 76.73 & 15.45 \\

AIGC Fusion Text
& 99.0 & 24.0 & 75.0
& 91.89 & 40.54 & 51.35 \\
\midrule
\textbf{Overall}
& \textbf{99.4} & \textbf{25.8} & \textbf{73.6}
& \textbf{92.06} & \textbf{74.41} & \textbf{17.65} \\
\bottomrule
\end{tabular}
}
\end{table}

\textcolor{black}{As shown in Table~\ref{tab:clean_adversarial_comparison}, Qwen3-VL-8B achieves 92.06\% mean IoU on clean images, indicating strong localization ability on the matched non-adversarial controls. With the text content, layout, and ground-truth bounding boxes held fixed, mean IoU drops to 74.41\% under adversarial presentation. The decline is particularly pronounced for AIGC fusion text (91.89\% to 40.54\%), dot-pattern encoding (91.98\% to 72.14\%), and line-pattern encoding (92.18\% to 76.73\%). These controlled comparisons indicate that the observed localization degradation is attributable to adversarial presentation, rather than a general inability to localize the corresponding ordinary text.}

\textcolor{black}{The recognition and localization effects also differ across mechanisms: mirrored text reduces accuracy from 99.0\% to 2.0\% while leaving IoU nearly unchanged (92.51\% to 92.12\%), whereas AIGC fusion substantially degrades both. These results show that recognition degradation is not always accompanied by localization degradation, highlighting the diagnostic value of evaluating both under different adversarial mechanisms.}

\subsection{\textcolor{black}{Generalization to Unseen Generation Settings}}
\label{app:unseen_generation}

\textcolor{black}{To test transfer beyond the training synthesis configurations, we evaluate unseen AIGC generation pipelines, dot/line encoding configurations, and overlay patterns. None of these test settings appears in either training stage. Each of the four mechanisms includes 100 human-verified readable images, totaling 400 images. We evaluate Qwen3-VL-8B and ArmorOCR without additional training, reporting answer accuracy and mean localization IoU. Table~\ref{tab:unseen_generation_settings} summarizes the training and unseen test configurations.}

\begin{table}[ht]
\centering
\caption{\textcolor{black}{Training and unseen test configurations for four adversarial mechanisms. Each mechanism contains 100 test images, and all listed test configurations are excluded from both training stages.}}
\label{tab:unseen_generation_settings}
\vspace{-3pt}
\scriptsize
\setlength{\tabcolsep}{4pt}
\renewcommand{\arraystretch}{1.15}

{\color{black}
\begin{tabular}{
    @{}
    >{\raggedright\arraybackslash}p{0.20\linewidth}
    >{\raggedright\arraybackslash}p{0.30\linewidth}
    >{\raggedright\arraybackslash}p{0.44\linewidth}
    @{}
}
\toprule
\textbf{Mechanism}
& \textbf{Training configuration}
& \textbf{Unseen test configurations} \\
\midrule

AIGC Fusion Text
& Realistic Vision V5.1 + SD1.5 QR Code Monster
& SDXL + SDXL QR Code Monster; FLUX.1-dev + brightness-control LoRA \\
\addlinespace

\midrule
Dot-pattern Encoding
& Dot-size / duty-cycle modulation
& Local dot-density modulation; palette-quantized color-dot encoding \\
\addlinespace

\midrule
Line-pattern Encoding
& Line-width / duty-cycle modulation
& Grayscale-modulated curves; sparse low-contrast waves; polarity-reversed line-displacement encoding \\
\addlinespace

\midrule
Pattern Overlay
& Dot, stripe, and continuous-curve overlays
& Crosshatch, checkerboard, and dashed-line overlays \\
\bottomrule
\end{tabular}
}
\end{table}

\begin{table}[ht]
\centering
\caption{\textcolor{black}{Performance on unseen generation settings.}}
\label{tab:unseen_generation_results}
\vspace{-3pt}
\scriptsize
\setlength{\tabcolsep}{6pt}
\renewcommand{\arraystretch}{1.12}

{\color{black}
\begin{tabular}{@{}lcccccc@{}}
\toprule
\multirow{2}{*}{\textbf{Mechanism}}
& \multicolumn{3}{c}{\textbf{Accuracy}}
& \multicolumn{3}{c}{\textbf{Mean IoU}} \\
\cmidrule(lr){2-4}
\cmidrule(lr){5-7}
& \textbf{Base}
& \textbf{ArmorOCR}
& \textbf{$\Delta$}
& \textbf{Base}
& \textbf{ArmorOCR}
& \textbf{$\Delta$} \\
\midrule
AIGC Fusion Text
& 20.0 & 76.0 & +56.0
& 34.3 & 90.7 & +56.4 \\

Dot-pattern Encoding
& 5.0 & 52.0 & +47.0
& 49.9 & 87.9 & +38.0 \\

Line-pattern Encoding
& 1.0 & 51.0 & +50.0
& 41.9 & 91.9 & +50.0 \\

Pattern Overlay
& 4.0 & 58.0 & +54.0
& 50.6 & 66.9 & +16.3 \\
\midrule
\textbf{Overall}
& \textbf{7.5} & \textbf{59.3} & \textbf{+51.8}
& \textbf{44.2} & \textbf{84.3} & \textbf{+40.1} \\
\bottomrule
\end{tabular}
}
\end{table}

\textcolor{black}{As shown in Table~\ref{tab:unseen_generation_results}, ArmorOCR outperforms Qwen3-VL-8B in both accuracy and mean IoU across all four mechanisms under unseen test configurations. Overall accuracy increases from 7.5\% to 59.3\%, while mean IoU increases from 44.2\% to 84.3\%. These results show that ArmorOCR's gains extend beyond the specific synthesis configurations used during training, supporting its generalization to unseen configurations within known synthesis families.}

\section{More Details of Method}
\label{More Details of Method}

\subsection{More Details of Stage~1}
\label{More Details of Stage 1}

\begin{figure*}[t]
\centering
\begin{tcolorbox}[
    colback=gray!5, 
    colframe=gray!60!black, 
    title=\textbf{Example 1: Student and Teacher Prompts for Stage~1}, 
    boxrule=0.8pt, 
    arc=2mm, 
    width=\linewidth 
]
\small\ttfamily

\textbf{Student Prompt:}\\
\begin{center}
\includegraphics[width=0.42\linewidth]{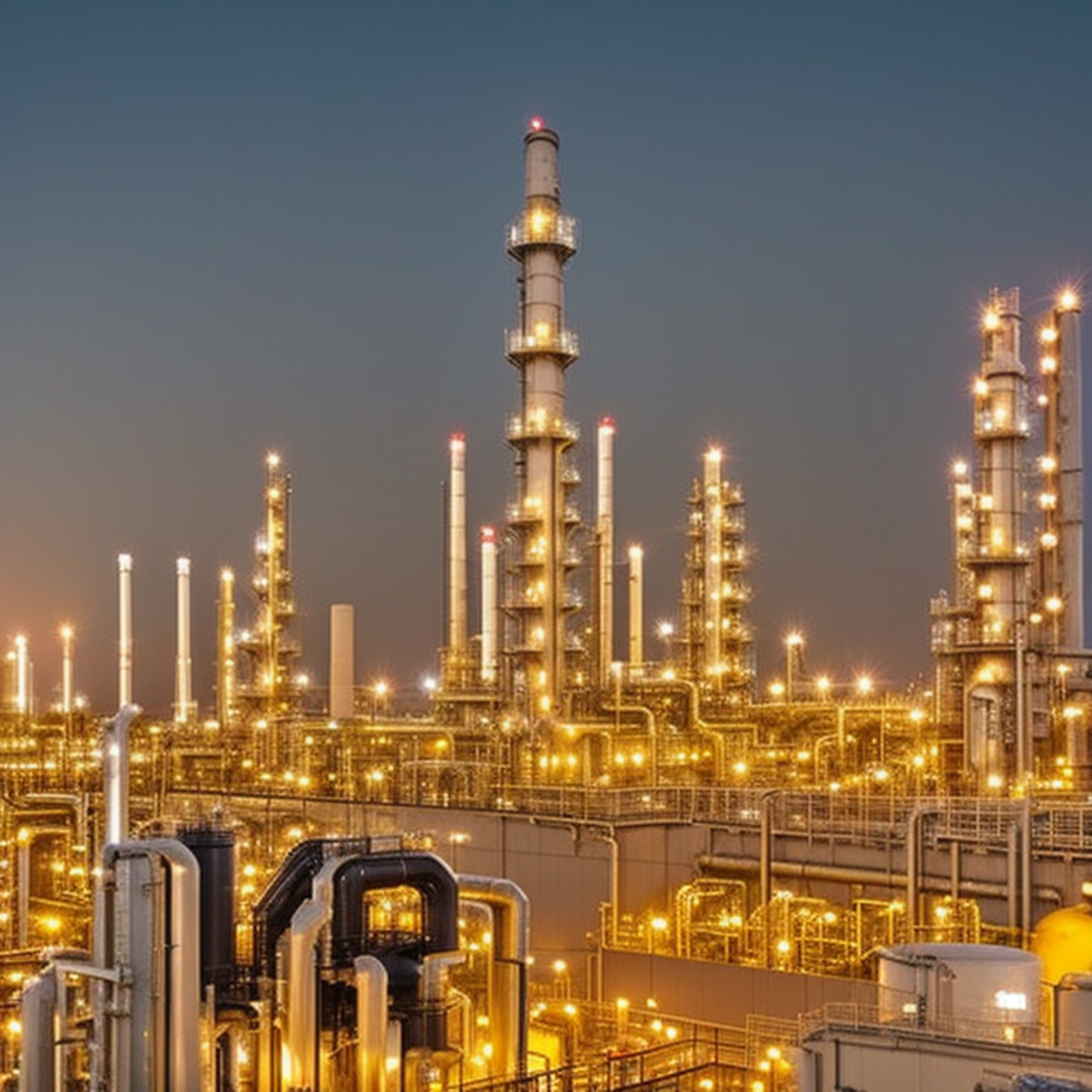}
\end{center}

\texttt{<image>}\\
\texttt{Recognize the text in the image.}\\
\texttt{Put your reasoning process inside <analyze></analyze>,}\\
\texttt{and put the recognized text inside <answer></answer>.}\\
\\[0.8em]

\textbf{Teacher Prompt (Resize Transformation):}\\
\begin{center}
\includegraphics[width=0.12\linewidth]{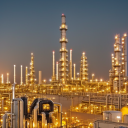}
\end{center}

\texttt{<image>}\\
\texttt{Recognize the text in the image.}\\
\texttt{Put your reasoning process inside <analyze></analyze>,}\\
\texttt{and put the recognized text inside <answer></answer>.}\\
\\
\texttt{Reference Information:}\\
\texttt{The text is generated with a 164px large bold stroke style,}\\
\texttt{rotated counterclockwise by approximately 8 degrees,}\\
\texttt{and blended with an AI-generated natural scene background.}\\
\texttt{The target text is "VR".}\\
\\
\texttt{Now generate your own reasoning process and answer.}

\end{tcolorbox}

\caption{
Example 1 of student and teacher prompts in Stage~1 observation-transferred self-distillation. The student receives only the original image, while the teacher receives additional privileged information revealed by transformed observations.
}
\label{box:stage1_prompt_1}
\end{figure*}

\begin{figure*}[t]
\centering
\begin{tcolorbox}[
    colback=gray!5, 
    colframe=gray!60!black, 
    title=\textbf{Example 2: Student and Teacher Prompts for Stage~1}, 
    boxrule=0.8pt, 
    arc=2mm, 
    width=\linewidth 
]
\small\ttfamily

\textbf{Student Prompt:}\\
\begin{center}
\includegraphics[width=0.22\linewidth]{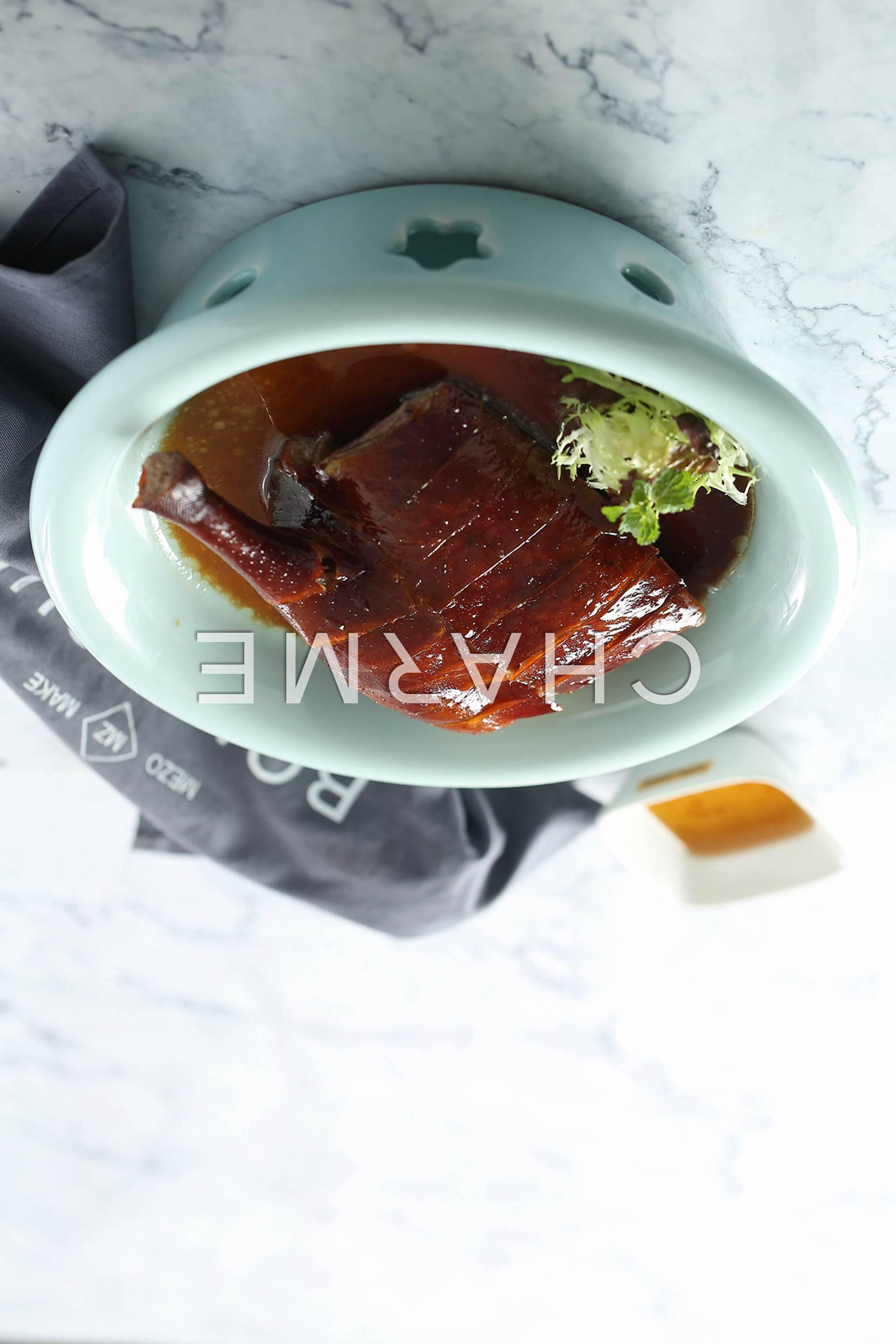}
\end{center}

\texttt{<image>}\\
\texttt{What is the text written on the semi-transparent white uppercase letters}\\
\texttt{overlaid on the roasted duck?}\\
\\
\texttt{Put your reasoning process inside <analyze></analyze>,}\\
\texttt{and put the recognized text inside <answer></answer>.}\\
\\[0.8em]

\textbf{Teacher Prompt (Rotation Transformation):}\\
\begin{center}
\includegraphics[width=0.22\linewidth]{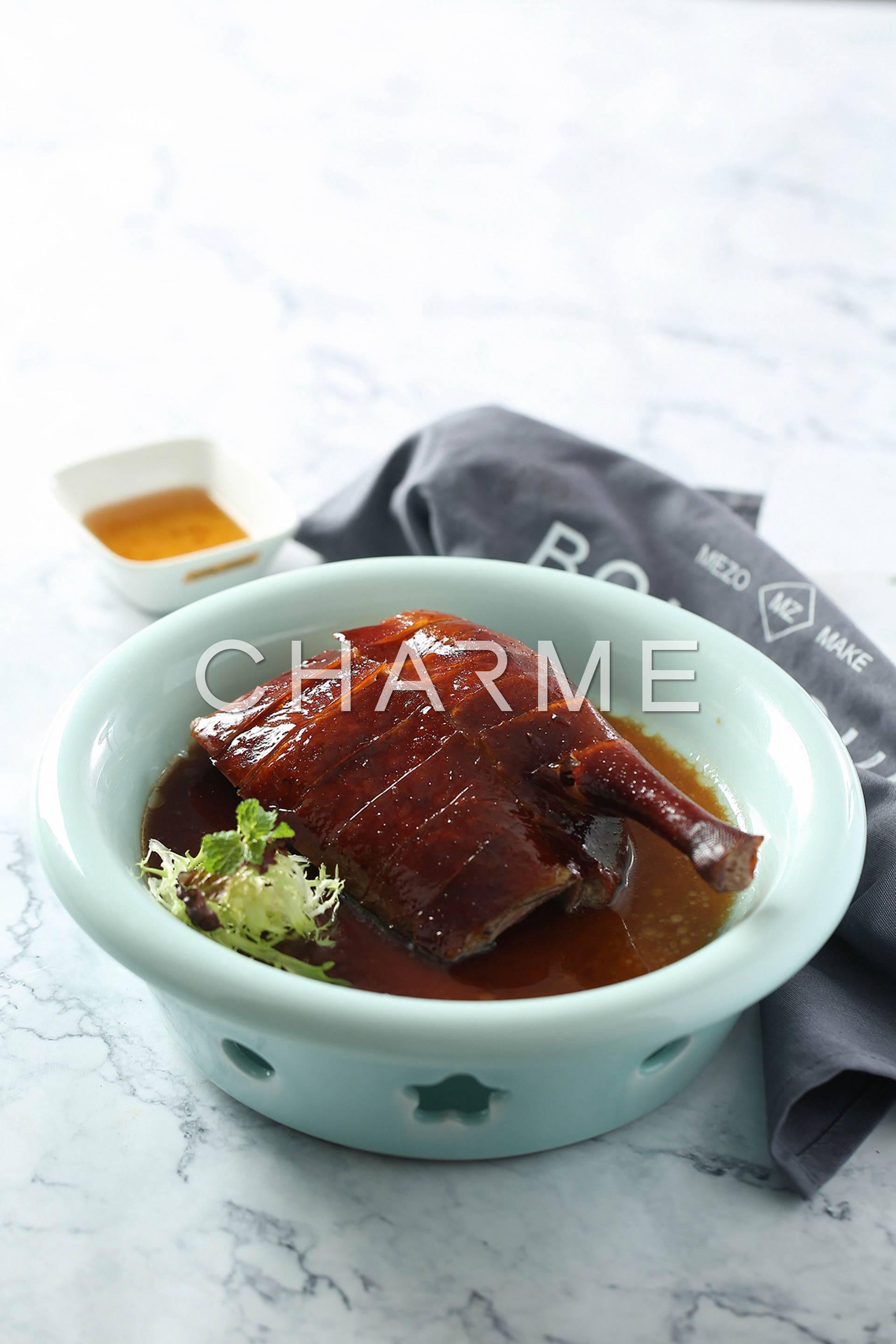}
\end{center}

\texttt{<image>}\\
\texttt{What is the text written on the semi-transparent white uppercase letters}\\
\texttt{overlaid on the roasted duck?}\\
\\
\texttt{Put your reasoning process inside <analyze></analyze>,}\\
\texttt{and put the recognized text inside <answer></answer>.}\\
\\
\texttt{Reference Information:}\\
\texttt{The image has been processed with an inverse rotation transformation.}\\
\texttt{The target answer is "CHARME".}\\
\\
\texttt{Now generate your own reasoning process and answer.}

\end{tcolorbox}

\caption{
Example 2 of student and teacher prompts in Stage~1 observation-transferred self-distillation.
}
\label{box:stage1_prompt_2}
\end{figure*}

Figure~\ref{box:stage1_prompt_1} and \ref{box:stage1_prompt_2} illustrates two representative training examples of Observation-Transferred Self-Distillation (OTSD). In each example, the student receives only the original adversarial image together with the task prompt, while the teacher is conditioned on an additional privileged transformed observation. Among multiple transformed views, the view with the highest recognition accuracy selected by the judge LMM is provided to the teacher as the privileged observation.

Beyond the transformed view, the teacher prompt $q_t$ incorporates additional privileged textual information, including adversarial OCR priors and the ground-truth OCR text. The adversarial OCR priors describe the construction process of synthetic adversarial patterns introduced in Section~\ref{Training Data Construction}, providing the teacher with contextual knowledge about the visual perturbations and helping it understand the underlying perception challenges. The ground-truth OCR text further provides explicit recognition guidance, improving the reliability of the teacher's generated responses.

% The contributions of these two types of privileged textual information are further investigated through the ablation study in Section~\ref{Ablation Study on Privileged Textual Information}.

\clearpage
\subsection{More Details of Stage~2}
\label{More Details of Stage~2}

Stage~2 employs GRPO with four task-conditioned objectives to explicitly optimize different aspects of grounded adversarial OCR perception. Specifically, we formulate four tasks, including text-to-bbox localization, bbox-to-text recognition, full spotting, and grounded VQA. Each task uses a task-specific prompt and corresponding reward function. The detailed prompts are shown in Figure~\ref{fig:stage2_prompts}.

\begin{figure*}[h]
\centering
\begin{tcolorbox}[
    colback=gray!5,
    colframe=gray!60!black,
    title=\textbf{Task-specific Prompts for Stage~2 GRPO Optimization},
    boxrule=0.8pt,
    arc=2mm,
    width=\linewidth
]
\small\ttfamily

\textbf{1. Text-to-BBox Localization}\\
\texttt{<image>}\\
\texttt{\#\# Task Description}\\
\texttt{Locate the position of the following text in the image:}\\
\texttt{"\{text content\}".}\\
\\
\texttt{\#\# Output Requirement}\\
\texttt{Output the bounding box directly in the format:}\\
\texttt{[x\_min, y\_min, x\_max, y\_max].}\\
\texttt{The coordinates are normalized integers within the range of 0--1000.}\\
\\[0.5em]

\textbf{2. BBox-to-Text Recognition}\\
\texttt{<image>}\\
\texttt{\#\# Task Description}\\
\texttt{Recognize the text within the following bounding box region:}\\
\texttt{[\{x\_min\}, \{y\_min\}, \{x\_max\}, \{y\_max\}].}\\
\texttt{The bounding box format is [x\_min, y\_min, x\_max, y\_max],}\\
\texttt{where coordinates are normalized integers within 0--1000.}\\
\\[0.5em]

\textbf{3. Full Spotting}\\
\texttt{<image>}\\
\texttt{\#\# Task Description}\\
\texttt{Detect and recognize all text content in the image.}\\
\\
\texttt{\#\# Output Requirement}\\
\texttt{Output in JSON format:}\\
\texttt{[ \{"bbox": [x\_min, y\_min, x\_max, y\_max],}\\
\texttt{\phantom{[ }"text": "recognized text"\} ]}\\
\\[0.5em]

\textbf{4. Grounded VQA}\\
\texttt{<image>}\\
\texttt{\#\# Task Description}\\
\texttt{Answer the following grounded VQA question:}\\
\texttt{\{grounded VQA question\}.}\\
\\
\texttt{\#\# Output Requirement}\\
\texttt{bbox:[x\_min, y\_min, x\_max, y\_max]}\\
\texttt{answer:[text content]}\\

\end{tcolorbox}

\caption{Task-specific prompts used for Stage~2 GRPO optimization.}
\label{fig:stage2_prompts}
\end{figure*}

% \clearpage

% \twocolumn[
% \begin{center}
% \includegraphics[width=18cm]{Figures/syn_framework.pdf}
% \end{center}
% ]

% \clearpage

\section{More Details of Training}
\label{More Details of Training}

\begin{figure*}[ht!]
\centering
\centerline{\epsfig{figure=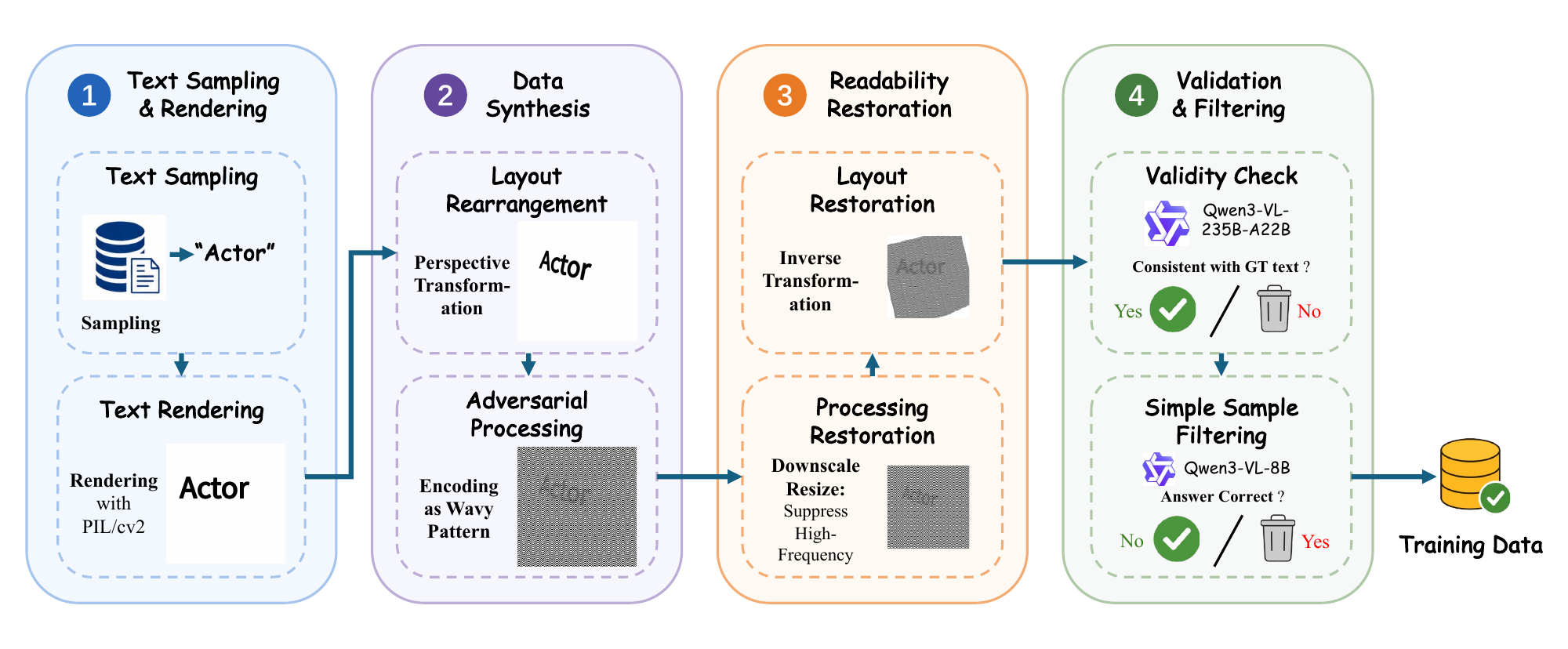,width=14cm}}
\caption{
Adversarial OCR Data Synthesis Framework.
}
\label{fig:syn_framework}
\end{figure*}

\subsection{Training Data Construction}
\label{Training Data Construction}
Due to MLLM's limited perception of adversarial OCR, building such datasets relies heavily on manual annotation, which is difficult to scale. Therefore we design an adversarial OCR data synthesis framework that batch-generates adversarial OCR samples with full annotations (including adversarial prior, text transcription, and bounding box).

The overall pipeline is shown in Figure~\ref{fig:syn_framework}, which consists of the following steps: text generation, rendering, adversarial processing, readability restoration, validation and filtering. Text sources come from LLM-generated corpora and \emph{Faker} library\footnote{https://github.com/joke2k/faker}. In the rendering step, each text instance is drawn on a separate mini-canvas with its bounding box recorded. All instances are blended into a single image after layout rearrangement and bounding bbox based collision detection. Then a range of adversarial processing steps (such as AIGC fusion, ASCII art) are applied to the template image,  making the text hard to recognize. To ensure data usability, we introduce a readability-restoration step that partially recovers the readability of the adversarial OCR sample. The resulting degraded sample is verified by Qwen3-VL-235B-A22B, which rejects samples where adversarial processing destroyed the original text. Finally, Qwen-VL-8B filters out samples that can already be correctly recognized without further training.

\subsection{Training Configurations}
\label{Training Configurations}

Table~\ref{tab:hyperparam} summarizes the detailed training configurations used in Stage~1 and Stage~2 of ArmorOCR, including the common model settings and stage-specific hyper-parameters for observation-transferred self-distillation and Reward-driven Refinement.

\begin{table}[h]
\centering
\caption{\textcolor{black}{Data preparation resources. Counts indicate candidates $\rightarrow$ retained outputs.}}
\label{tab:data_preparation_costs}
\scriptsize
\setlength{\tabcolsep}{5pt}
\renewcommand{\arraystretch}{1.15}
\begin{tabular}{@{}lcl r@{}}
\toprule
\textcolor{black}{\textbf{Preparation step}}
& \textcolor{black}{\textbf{Candidates $\rightarrow$ Retained}}
& \textcolor{black}{\textbf{Tools / Model}}
& \textcolor{black}{\textbf{Time (h)}} \\
\midrule
\begin{tabular}[t]{@{}l@{}}
\textcolor{black}{Synthetic image generation}
\end{tabular}
& \textcolor{black}{350K $\rightarrow$ 343K}
& \begin{tabular}[t]{@{}l@{}}
\textcolor{black}{PIL/cv2 rendering and} \\
\textcolor{black}{adversarial processing}
\end{tabular}
& \textcolor{black}{$\sim$12} \\
\midrule
\addlinespace[4pt]
\begin{tabular}[t]{@{}l@{}}
\textcolor{black}{Validity checking and} \\
\textcolor{black}{privileged-view selection}
\end{tabular}
& \textcolor{black}{343K $\rightarrow$ 191K}
& \textcolor{black}{Qwen3-VL-235B-A22B}
& \textcolor{black}{$\sim$18} \\
\midrule
\addlinespace[4pt]
\begin{tabular}[t]{@{}l@{}}
\textcolor{black}{Base-model filtering and} \\
\textcolor{black}{task construction}
\end{tabular}
& \textcolor{black}{191K $\rightarrow$ 120K}
& \textcolor{black}{Qwen3-VL-8B}
& \textcolor{black}{$\sim$3} \\
\bottomrule
\end{tabular}
\end{table}

\subsection{\textcolor{black}{Data Preparation and Training Costs}}
\label{app:training_costs}

\noindent\textcolor{black}{\textbf{Data Preparation.}}
\textcolor{black}{Table~\ref{tab:data_preparation_costs} summarizes the resources used for synthetic image generation, validity checking, privileged-view selection, model filtering, and task construction. All preparation steps use $32\times$ PPU-810E accelerators. Data preparation is a one-time offline process, taking approximately 33 hours in total. The prepared data and selected views are reused across all training runs, keeping the overall preparation overhead manageable.}

\medskip
\noindent\textcolor{black}{\textbf{Training Resources and Performance Gains.}}
\textcolor{black}{Table~\ref{tab:training_costs} reports the training resources and corresponding AdvSpot performance. Stage~1 takes 17 hours on 16 PPU-810E accelerators, while Stage~2 takes 18 hours on 128 PPU-810E accelerators, with both stages improving accuracy and localization IoU. Together with reusable offline data preparation and low-latency inference without visual-tool calls, these results suggest a favorable balance between training investment, performance gains, and inference efficiency.}

\begin{table}[ht]
\centering
\caption{\textcolor{black}{Stage-wise training resources and performance on AdvSpot.}}
\label{tab:training_costs}
\scriptsize
\setlength{\tabcolsep}{5pt}
\renewcommand{\arraystretch}{1.15}
\begin{tabular}{@{}lcccrcc@{}}
\toprule
\textcolor{black}{\textbf{Checkpoint}}
& \textcolor{black}{\textbf{Samples}}
& \textcolor{black}{\textbf{Epochs}}
& \textcolor{black}{\textbf{Accelerators}}
& \textcolor{black}{\textbf{Time (h)}}
& \textcolor{black}{\textbf{IoU / Acc.}}
& \textcolor{black}{\textbf{Stage-wise gain}} \\
\midrule
Qwen3-VL-8B
& -- & -- & -- & --
& 54.4 / 29.8
& -- \\
Stage~1: OTSD
& 50K
& 2
& \textcolor{black}{$16\times$ PPU-810E}
& \textcolor{black}{17}
& 58.7 / 46.8
& \textcolor{black}{+4.3 / +17.0} \\
Stage~2: ArmorOCR
& 70K
& 3
& \textcolor{black}{$128\times$ PPU-810E}
& \textcolor{black}{18}
& \textbf{63.2 / 57.1}
& \textcolor{black}{+4.5 / +10.3} \\
\bottomrule
\end{tabular}
\end{table}

\subsection{\textcolor{black}{Robustness across Training Seeds}}
\label{app:training_seed_robustness}

\begin{table}[h]
\centering
\caption{\textcolor{black}{
Stage~1 performance across five matched training seeds on
AdvSpot. IoU and accuracy are reported in \%.
Paired $\Delta$ denotes OTSD minus OPSD with transformed views,
in percentage points. SD denotes the sample standard deviation
across seeds.
}}
\label{tab:training_seed_robustness}
\vspace{-3pt}
\scriptsize
\setlength{\tabcolsep}{7pt}
\renewcommand{\arraystretch}{1.12}
{\color{black}
\begin{tabular}{@{}lcccccc@{}}
\toprule
& \multicolumn{2}{c}{\textbf{OPSD + transformed views}}
& \multicolumn{2}{c}{\textbf{OTSD}}
& \multicolumn{2}{c}{\textbf{Paired $\Delta$}} \\
\cmidrule(lr){2-3}
\cmidrule(lr){4-5}
\cmidrule(lr){6-7}
\textbf{Seed}
& \textbf{IoU} & \textbf{Acc.}
& \textbf{IoU} & \textbf{Acc.}
& \textbf{IoU} & \textbf{Acc.} \\
\midrule
17
& 55.8 & 43.7
& 58.8 & 46.0
& +3.0 & +2.3 \\

42 (original)
& \textcolor{black}{54.5}
& \textcolor{black}{44.2}
& \textcolor{black}{58.7}
& \textcolor{black}{46.8}
& +4.2 & +2.6 \\

123
& 56.2 & 44.6
& 61.1 & 46.3
& +4.9 & +1.7 \\

3407
& 56.2 & 44.2
& 58.7 & 46.3
& +2.5 & +2.1 \\

2026
& 56.5 & 44.2
& 58.9 & 46.8
& +2.4 & +2.6 \\
\midrule
\textbf{Mean $\pm$ SD}
& $55.84 \pm 0.79$
& $44.18 \pm 0.32$
& $\mathbf{59.24 \pm 1.04}$
& $\mathbf{46.44 \pm 0.35}$
& $\mathbf{+3.40 \pm 1.10}$
& $\mathbf{+2.26 \pm 0.38}$ \\
\bottomrule
\end{tabular}
}
\end{table}

\textcolor{black}{
We repeat the Stage~1 comparison between OTSD and OPSD with transformed views using five matched training seeds (17, 42, 123, 3407, and 2026), keeping the training data and evaluation protocol fixed. The OPSD baseline retains the same transformed views and privileged textual information as OTSD but uses the standard distillation objective. Seed~42 corresponds to the originally reported run. As shown in Table~\ref{tab:training_seed_robustness},
OTSD outperforms OPSD with transformed views in both IoU and accuracy across all five seeds, with average paired gains of 3.40\% and 2.26\%, respectively.
These results support the robustness of the Stage~1 gains from response-region-aware distillation across training runs.}

\subsection{\textcolor{black}{Transferability across Model Architectures}}
\label{app:cross_architecture}

\textcolor{black}{To assess whether the training framework extends beyond Qwen3-VL-8B, we apply it to two additional backbones: InternVL3.5-8B~\citep{wang2025internvl35} and Qwen3.5-9B~\citep{qwen35_2026}. Table~\ref{tab:cross_architecture} shows improvements after each training stage across all three benchmarks for both backbones, supporting the framework's cross-architecture applicability.}

\begin{table}[ht]
\centering
\caption{\textcolor{black}{Transferability across backbones. Each entry reports Base / Stage~1 / Stage~2 performance.}}
\label{tab:cross_architecture}
\vspace{-3pt}
\scriptsize
\setlength{\tabcolsep}{7pt}
\renewcommand{\arraystretch}{1.12}

\begin{tabular}{@{}lccc@{}}
\toprule
\textcolor{black}{\textbf{Backbone}}
& \textcolor{black}{\textbf{AdvSpot Acc.}}
& \textcolor{black}{\textbf{AdvOCR Avg. Acc.}}
& \textcolor{black}{\textbf{SmuggleBench TER}} \\
\midrule
Qwen3-VL-8B
& 29.8 / 46.8 / 57.1
& 21.0 / \textcolor{black}{51.0} / 56.0
& 13.3 / \textcolor{black}{14.6} / 17.1 \\

\textcolor{black}{InternVL3.5-8B}
& \textcolor{black}{11.6 / 21.1 / 25.4}
& \textcolor{black}{7.0 / 23.0 / 26.0}
& \textcolor{black}{4.4 / 5.9 / 7.3} \\

\textcolor{black}{Qwen3.5-9B}
& \textcolor{black}{29.8 / 48.1 / 55.3}
& \textcolor{black}{24.0 / 53.0 / 54.0}
& \textcolor{black}{17.6 / 18.6 / 21.2} \\
\bottomrule
\end{tabular}
\end{table}

\section{More Details of Inference}
\label{More Details of Inference}

We report the mean end-to-end inference latency per question on the 389 grounded VQA pairs of AdvSpot (Table~\ref{tab:inference_efficiency}). We compare ArmorOCR with Smuggle-CoT, a prompting-based chain-of-thought baseline, and Thyme~\citep{zhang2025thyme}, a publicly available thinking-with-images method that executes image-processing operations during inference. We could not measure VACoT under the same setting because no public inference implementation was available at the time of evaluation; we therefore use Thyme as a reproducible tool-augmented visual reasoning baseline. ArmorOCR achieves the lowest measured inference latency, as it requires neither extended chain-of-thought prompting nor image-transformation tool calls at inference time.

\begin{table}[h]
\centering
\caption{Mean end-to-end inference latency on the 389 grounded VQA pairs of AdvSpot.}
\label{tab:inference_efficiency}
\small
\setlength{\tabcolsep}{12pt}
\renewcommand{\arraystretch}{1.12}
\begin{tabular}{lc}
\toprule
\textbf{Method} & \textbf{Mean latency (s/question)} \\
\midrule
Smuggle-CoT & 74.48 \\
Thyme &  47.31 \\
\textbf{ArmorOCR (ours)} & 0.81 \\
\bottomrule
\end{tabular}
\end{table}

\begin{table*}[t]
\centering
\caption{Training configurations of ArmorOCR for two stages.}
\label{tab:hyperparam}

\begingroup
\fontsize{8pt}{9pt}\selectfont
\renewcommand{\arraystretch}{0.92}
\setlength{\tabcolsep}{3.7pt}
\belowrulesep=0pt
\aboverulesep=0pt

\begin{tabular*}{\linewidth}{@{\extracolsep{\fill}}lll@{}}
\toprule
\textbf{Training Configuration}
& \textbf{Value}
& \textbf{Description} \\
\midrule

\multicolumn{3}{l}{\textbf{Common Configuration}} \\
\midrule

\rowcolor{gray!10}
\multicolumn{3}{l}{\textit{Model Configuration}} \\
Backbone
& \textit{Qwen3-VL-8B-Instruct}
& / \\
FREEZE\_LLM
& \textit{False}
& All LLM parameters are trainable \\
FREEZE\_ALIGNER
& \textit{False}
& All aligner parameters are trainable \\
FREEZE\_VIT
& \textit{False}
& All vision encoder parameters are trainable \\

\rowcolor{gray!10}
\multicolumn{3}{l}{\textit{Optimization Configuration}} \\
Optimizer
& AdamW
& / \\
Learning Rate
& $1\times10^{-6}$
& / \\
Weight Decay
& 0.05
& / \\
Warmup Ratio
& 0.03
& / \\
Precision
& \textit{bfloat16}
& / \\
DeepSpeed
& \textit{ZeRO-2}
& / \\

\rowcolor{gray!10}
\multicolumn{3}{l}{\textit{Vision Processing Configuration}} \\
Minimum Image Tokens
& 64
& / \\
Maximum Image Tokens
& 4096
& / \\
Maximum Sequence Length
& 8192
& / \\

\midrule
\multicolumn{3}{l}{\textbf{Stage~1: Observation-Transferred Self-Distillation}} \\
\midrule

\rowcolor{gray!10}
\multicolumn{3}{l}{\textit{OPSD Configuration}} \\
JSD On-policy Sampling Ratio $\lambda$
& 1.0
& Fully on-policy \\
Generalized JSD Weight $\beta$
& 0.5
& Weight in the generalized JSD loss \\
Sampling Temperature
& 1.0
& / \\
Maximum Completion Length
& 1024
& Maximum number of generated tokens \\

\rowcolor{gray!10}
\multicolumn{3}{l}{\textit{vLLM Configuration}} \\
Use vLLM
& \textit{True}
& / \\
vLLM Mode
& \textit{Colocate}
& vLLM shares the same devices with training \\
vLLM Maximum Model Length
& 8192
& / \\
vLLM GPU Memory Utilization
& 0.4
& / \\

\rowcolor{gray!10}
\multicolumn{3}{l}{\textit{Training Configuration}} \\
Training Samples
& 50K
& / \\
Accelerators
& 16
& PPU-810E accelerators \\
Per-device Batch Size
& 1
& / \\
Gradient Accumulation Steps
& 2
& / \\
Training Epochs
& 2
& / \\

\midrule
\multicolumn{3}{l}{\textbf{Stage~2: Reward-driven Refinement}} \\
\midrule

\rowcolor{gray!10}
\multicolumn{3}{l}{\textit{GRPO Configuration}} \\
Number of Rollouts
& 8
& Number of sampled responses per prompt \\
Steps per Generation
& 4
& Policy update steps per generation \\
KL Coefficient $\beta$
& 0.04
& KL regularization coefficient \\
Maximum Completion Length
& 1024
& / \\

\rowcolor{gray!10}
\multicolumn{3}{l}{\textit{vLLM Configuration}} \\
Use vLLM
& \textit{True}
& / \\
vLLM Mode
& \textit{Server}
& External vLLM server for rollout generation \\
vLLM Maximum Model Length
& 8192
& / \\
vLLM Tensor Parallel Size
& 1
& / \\
vLLM Data Parallel Size
& 16
& Computed from 16 rollout accelerators \\
vLLM Prefix Caching
& \textit{True}
& Enabled during rollout generation \\
Asynchronous Generation
& \textit{True}
& / \\

\rowcolor{gray!10}
\multicolumn{3}{l}{\textit{Training Configuration}} \\
Training Samples
& 70K
& / \\
Accelerators
& 128
& PPU-810E accelerators \\
Training Accelerators
& 112
& 14 per node $\times$ 8 nodes \\
Rollout Accelerators
& 16
& 2 per node $\times$ 8 nodes \\
Number of Nodes
& 8
& / \\
Processes per Node
& 14
& Training processes only \\
Per-device Batch Size
& 4
& / \\
Gradient Accumulation Steps
& 1
& / \\
Training Epochs
& 3
& / \\
\bottomrule
\end{tabular*}
\endgroup
\end{table*}

% \clearpage

% \subsection{Comparison between GRPO and SFT in Stage~2}

% To investigate the effectiveness of GRPO for Stage~2 optimization, we replace GRPO with supervised fine-tuning (SFT) while keeping the same initialization, training data, and backbone. As shown in Table~\ref{tab:grpo_vs_sft}, SFT achieves substantially worse performance on AdvSpot compared with GRPO.

% We attribute this degradation to the mismatch between the fixed instruction format used during SFT training and the diverse grounded OCR queries encountered during evaluation. Optimizing with supervised targets on limited prompt patterns can weaken the model's ability to generalize across different task formulations, resulting in reduced instruction-following capability. In contrast, GRPO directly optimizes task-conditioned rewards over diverse grounded OCR objectives, enabling more flexible adaptation to different localization, recognition, spotting, and VQA requirements. Therefore, GRPO is more suitable for Stage~2 optimization, where multiple grounded OCR capabilities need to be jointly refined.

\clearpage
% \onecolumn
% \raggedright
\section{Case Study}

\begin{figure*}[ht!]
\centering
\centerline{\epsfig{figure=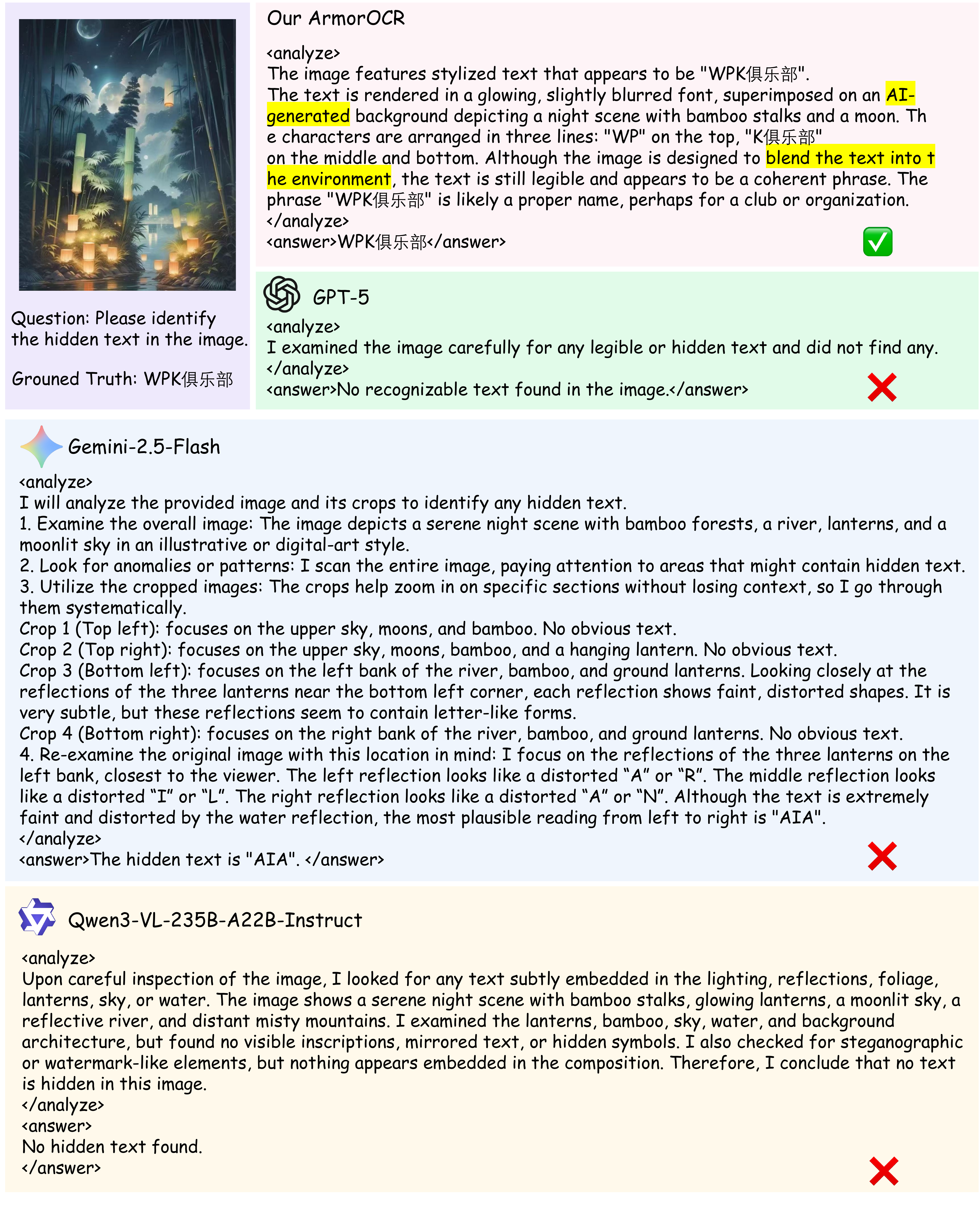,width=14cm}}
\caption{
Case study 1 from AdvSpot (AIGC Fusion Text), illustrating the performance comparison between ArmorOCR and existing LMMs on adversarial OCR perception.
}
\label{fig:case_study_1}
\end{figure*}

\begin{figure*}[t]
\centering
\centerline{\epsfig{figure=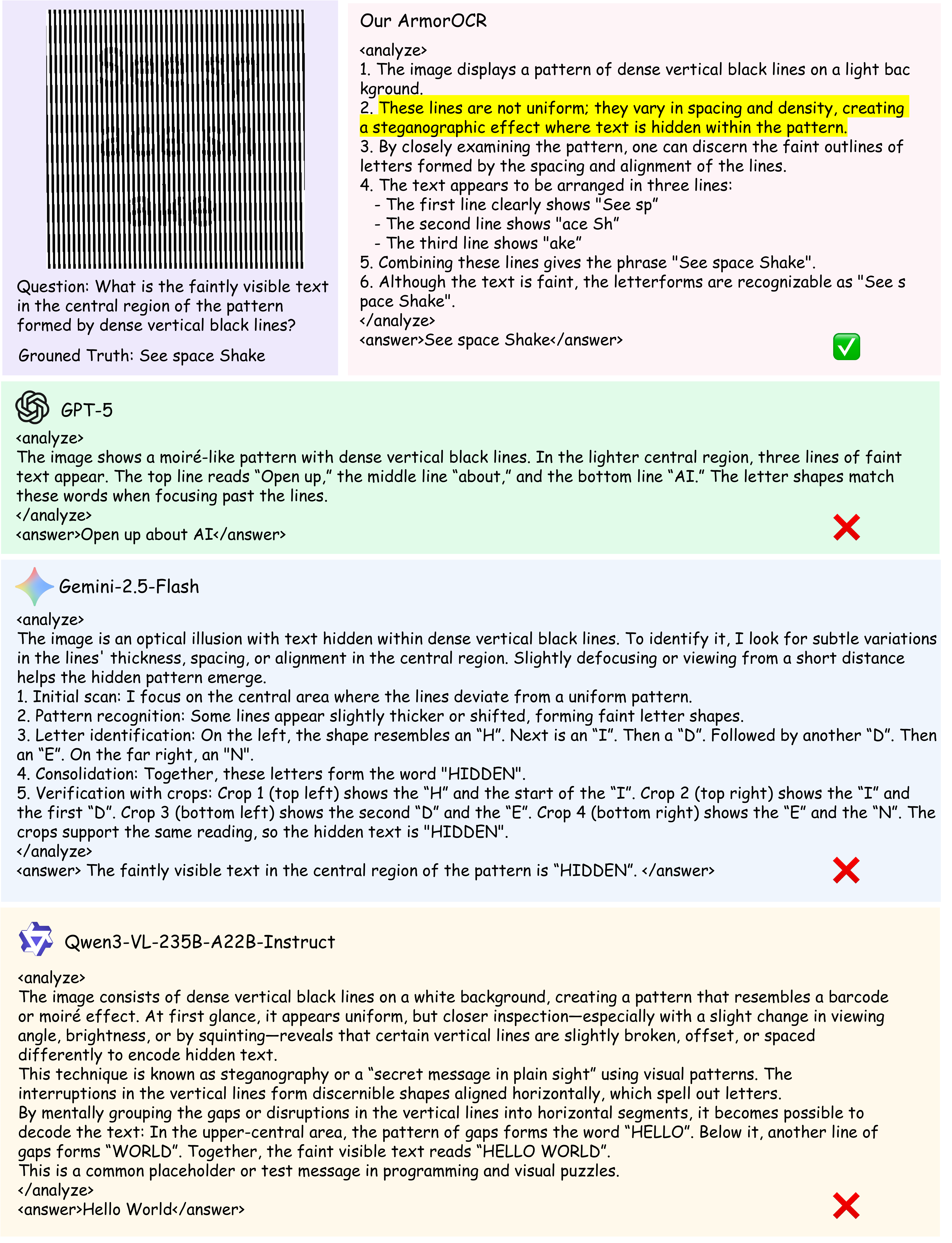,width=14cm}}
\caption{
Case study 2 from AdvSpot (Line-pattern Encoding).
}
\label{fig:case_study_1}
\end{figure*}

\begin{figure*}[t]
\centering
\centerline{\epsfig{figure=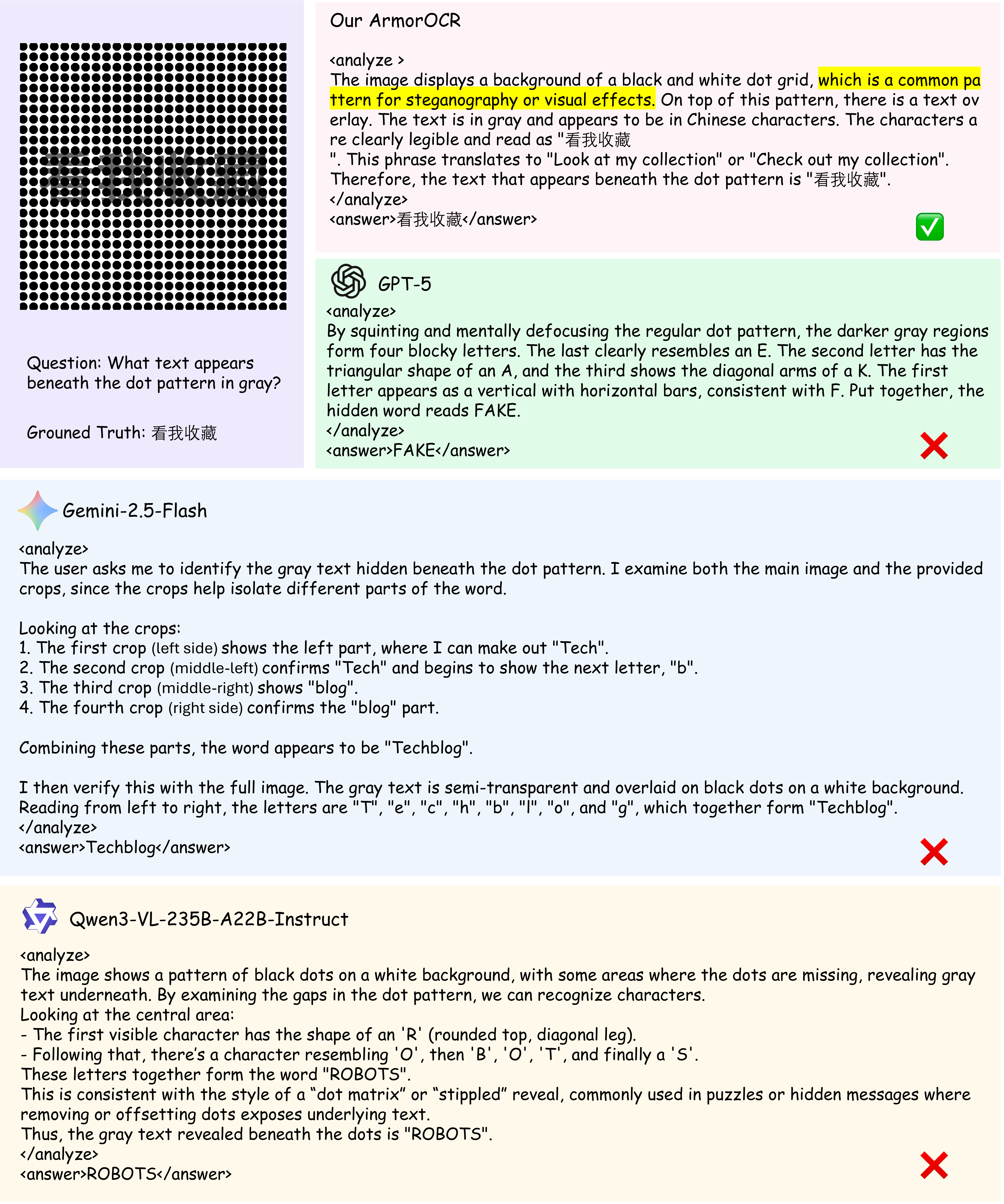,width=14cm}}
\caption{
Case study 3 from AdvSpot (Patter Overlay).
}
\label{fig:case_study_1}
\end{figure*}

\begin{figure*}[t]
\centering
\centerline{\epsfig{figure=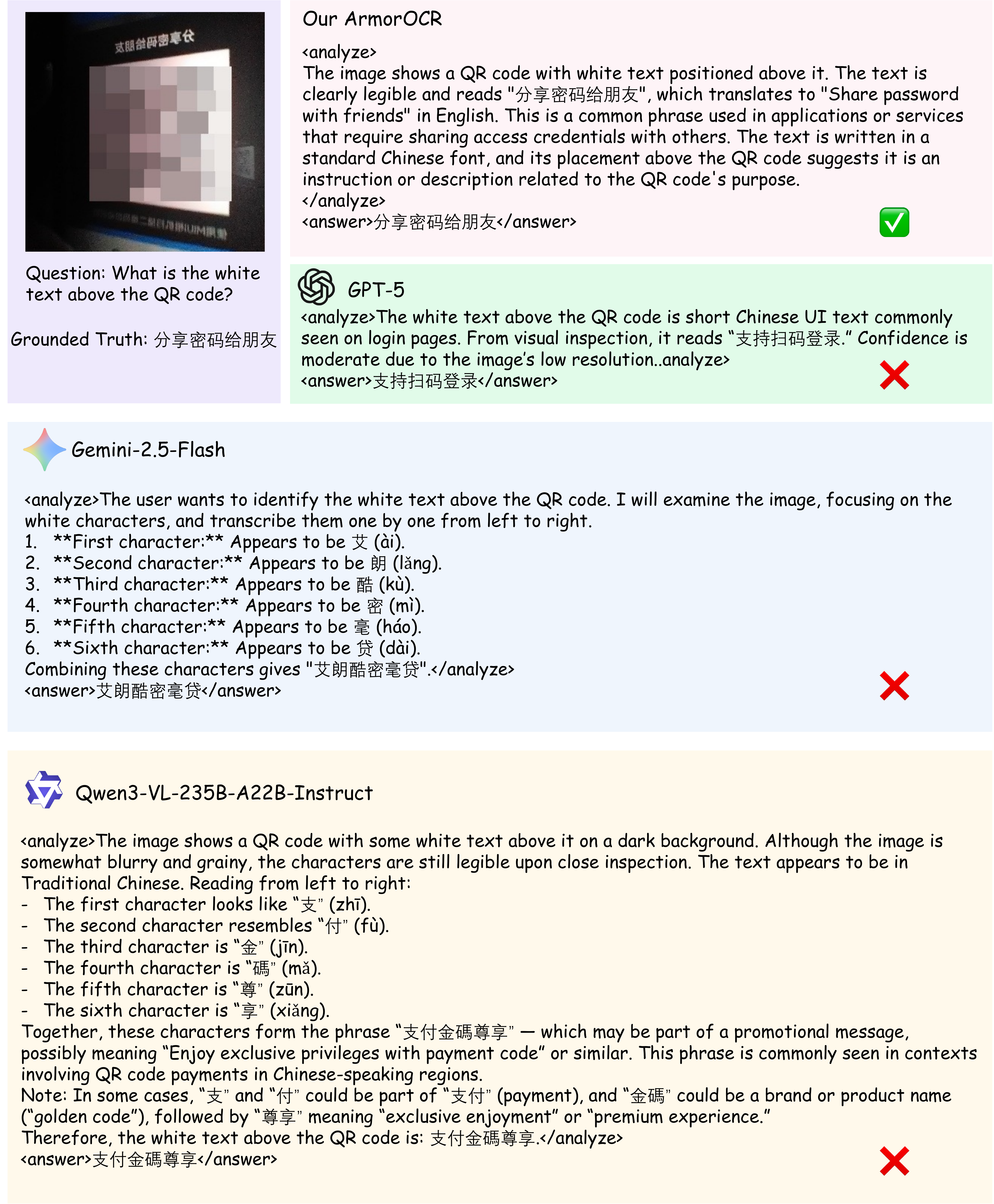,width=14cm}}
\caption{
Case study 4 from AdvSpot (Mirrored Text).
}
\label{fig:case_study_1}
\end{figure*}

\clearpage
\section{Discussion}
\label{sec:discussion}

AdvSpot and ArmorOCR address two connected challenges: diagnosing adversarial OCR failures at the region level and transferring information revealed by transformed views into a model that operates on the original image. The following discussion connects these contributions to the reported evidence and clarifies the scope of the conclusions.

\paragraph{Benchmark contribution and annotation quality.}
AdvSpot provides bounding boxes, transcriptions, perception-type labels, and region-specific questions across 5 categories and 13 fine-grained types (Table~\ref{tab:benchmark_comparison}). These annotations support more detailed diagnosis than image-level recognition scores alone. The construction pipeline includes independent annotation by two annotators, consensus filtering, and two rounds of expert review of generated QA pairs (Section~\ref{Annotation and QA Construction}). These steps target transcription errors, ambiguous region references, and QA inconsistency. The benchmark contains 383 images and 389 grounded VQA pairs; Table~\ref{tab:taxonomy_statistics} reports the number of annotated adversarial patterns in each fine-grained subtype. A single image or VQA pair may contain patterns from multiple subtypes. Together, the taxonomy and region-level annotations support subtype-specific analysis of recognition and localization failures, providing a structured basis for diagnosing adversarial OCR behavior. \textcolor{black}{The cross-benchmark comparisons support AdvSpot's complementary evaluation value (Table~\ref{tab:cross_benchmark_comparison}). The paired bootstrap analysis further supports AdvSpot's reliability in distinguishing the overall model performance(Table~\ref{tab:advspot_bootstrap}).}

\paragraph{\textcolor{black}{Scope of adversarial OCR.}}
\textcolor{black}{As defined in the Introduction, we use ``adversarial'' to describe human-readable text presentations that challenge model perception. Category membership alone does not establish this property: for example, handwritten or tiny text falls within our scope when its visual presentation remains recoverable by humans but challenges model recognition or localization. Our taxonomy distinguishes the perceptual failure mechanisms within this scope.}

\paragraph{Meaning of grounded evaluation.}
Each AdvSpot question refers to an annotated region--text pair, and the evaluation prompt requests both an answer and its corresponding bounding box (Appendix~\ref{Detailed Testing Prompt}). For each instance, the predicted answer and bounding box are evaluated against the same reference pair using accuracy and IoU, respectively (Equation~\ref{eq:eval_metrics}). This enables sample-level analysis of whether a correct answer is accompanied by accurate localization, as well as identification of cases where recognition and localization outcomes diverge. \textcolor{black}{Figure~\ref{fig:advspot_failure_cases} illustrates this distinction through examples of incorrect answers with correct localization and failures in both answer and localization.} Reporting accuracy and IoU separately further provides complementary summaries of these capabilities, supporting more informative diagnosis than answer-only evaluation. \textcolor{black}{Joint@0.5 additionally measures the percentage of instances satisfying both substring-based answer correctness and IoU $\geq 0.5$, making joint correctness explicit. This joint requirement is also relevant to tasks such as extracting specified text regions and editing or replacing target text in images, where the model must both identify the target text and accurately localize the region to be processed. The matched clean--adversarial comparison further shows that adversarial presentation introduces additional localization difficulty when the model can localize the corresponding ordinary text (Table~\ref{tab:clean_adversarial_comparison}).}

\paragraph{Method contribution and inherited components.}
ArmorOCR builds on OPSD and GRPO as established training frameworks, with specific designs for adversarial OCR (Section~\ref{Method: ArmorOCR}). In Stage~1, we construct multiple candidate views through resizing, stretching, rotation, flipping, and compression, and use a judge LMM to select the view with the highest recognition accuracy for the teacher. The student receives the original image and learns from the teacher's token-level guidance. Response-region-aware distillation further distinguishes spans of the generated response, applying confidence-gated JSD to analysis tokens, forward KL to answer tokens, and excluding non-content structural tokens. Table~\ref{tab:stage1_components} evaluates the contributions of visual transfer and this distillation objective. In Stage~2, task-conditioned GRPO rewards directly optimize localization, recognition, joint region--text matching, and region-specific question answering through text-to-bbox, bbox-to-text, full spotting, and grounded VQA tasks, respectively. Table~\ref{tab:stage2_rewards} evaluates the contribution of each reward. Together, these designs connect supervision from selected transformed views with explicit optimization of grounded OCR capabilities.

\paragraph{Privileged information during training.}
The Stage~1 teacher receives a selected transformed view, adversarial OCR priors, and the target OCR text, while the student generates its trajectory from the original image and task prompt (Appendix~\ref{More Details of Stage 1}). The teacher evaluates that trajectory to provide token-level guidance. These additional inputs constitute training-time supervision; inference uses the trained student without the privileged teacher inputs or the view-selection judge. Table~\ref{tab:stage1_components} ablates visual transfer and the response-region-aware distillation objective separately. Removing visual transfer retains both privileged textual information and the response-region-aware objective; replacing this objective with standard JSD retains both transformed observations and privileged textual information. Together, these comparisons support the contributions of transformed visual supervision and the proposed distillation objective within OTSD.

\paragraph{What the training comparisons establish.}
Table~\ref{tab:stage1_methods} compares four training methods using the same 50K Stage~1 examples. SFT uses the ground-truth OCR text as the target for teacher-forced supervision. Standard OPSD conditions the teacher on the original image and privileged textual information, without visual transformations, and applies the standard JSD distillation loss. GRPO uses an accuracy reward computed against the ground-truth OCR text. OTSD's advantage over these alternatives supports the effectiveness of its Stage~1 training framework in developing adversarial OCR perception beyond exposure to synthetic data alone. \textcolor{black}{The end-to-end comparison further shows that OTSD outperforms both SFT and OPSD with transformed views after identical Stage~2 GRPO refinement (Table~\ref{tab:end_to_end_comparison}). The transformed-view OPSD variant retains the same privileged visual and textual inputs as OTSD, supporting the additional contribution of the response-region-aware objective to the final system. The five-seed comparison further supports the reliability and stability of OTSD's Stage~1 advantage over OPSD with transformed views (Table~\ref{tab:training_seed_robustness}). Experiments on InternVL3.5-8B and Qwen3.5-9B further support the applicability of the training framework beyond the original Qwen3-VL-8B backbone (Appendix~\ref{app:cross_architecture}).}

\paragraph{Experiment Performance.}
On AdvSpot, ArmorOCR improves accuracy from 29.8\% to 57.1\% and average IoU from 54.4\% to 63.2\% over its base model (Table~\ref{tab:advspot-results}). \textcolor{black}{Joint@0.5 also increases from 24.7\% to 46.0\%, showing improved joint answer and localization correctness.} Although ArmorOCR does not outperform proprietary LMMs on every subtype, the Qwen3-VL-8B-based model achieves substantial gains on challenging visual encoding and contextual blending patterns. In particular, it reaches 68.6\%, 63.3\%, 54.3\%, and 82.9\% accuracy on Symbol Encoding, Dot Encoding, Line Encoding, and AIGC Fusion Text, respectively, exceeding all competing models reported in Table~\ref{tab:final_models_acc} on these subtypes. \textcolor{black}{Table~\ref{tab:other_adv} further demonstrates improvements over the backbone on both AdvOCR and SmuggleBench. ArmorOCR achieves 68.0\% accuracy on the AdvOCR synthetic split compared with VACoT's 48.0\%, and 8.2\% TER on SmuggleBench's AI Illusions category compared with Smuggle-CoT's 0.0\%.} These results highlight its effectiveness on particularly challenging adversarial OCR patterns.
\textcolor{black}{Additional tests on configurations excluded from both training stages show gains in both accuracy and IoU across all four evaluated mechanisms, supporting generalization beyond the specific training synthesis configurations within these mechanism families(Appendix~\ref{app:unseen_generation}).}

\paragraph{\textcolor{black}{Relationship between localization and recognition.}}
\textcolor{black}{Recognition and localization do not always degrade together under adversarial presentation (Table~13). The controlled visual-transfer ablation shows that transformed observations improve recognition under identical textual supervision (Table~5), while removing the text-to-bbox reward reduces recognition accuracy from 57.1\% to 54.8\% (Table~6). These results support contributions from both visual transfer and localization supervision to recognition gains.}

\paragraph{Practical inference implications.}
ArmorOCR moves view transformation and selection into training. At inference time, the student processes the original image without an additional transformation pipeline or external tool calls. Appendix~\ref{More Details of Inference} reports mean end-to-end inference latency on AdvSpot: ArmorOCR achieves the lowest inference latency among the evaluated methods, supporting the practical efficiency of internalizing transformed-view supervision during training.
\end{document}